\documentclass{article}
\ifdefined\pdfinfoomitdate\pdfinfoomitdate=1\fi
\ifdefined\pdftrailerid\pdftrailerid{}\fi
\ifdefined\pdfsuppressptexinfo\pdfsuppressptexinfo=-1\fi
\ifdefined{/Producer (pdfTeX)}\fi
\usepackage{iclr2027_conference,times}
\iclrfinalcopy
\usepackage[utf8]{inputenc}
\usepackage[T1]{fontenc}
\usepackage[hidelinks]{hyperref}
\usepackage{url}
\usepackage{amsmath,amssymb,amsfonts}
\usepackage{graphicx}
\usepackage{booktabs}
\usepackage{array}
\usepackage{multirow}
\usepackage{xcolor}
\usepackage{natbib}
\newcommand{\todo}[1]{\unskip}
\newcommand{\num}[1]{#1}
\newif\ifreinflowdata
\reinflowdatatrue

\title{Uncertainty-Gated Exploration Noise Suppresses Task Collapse in Online RL Fine-Tuning of a Flow-Matching Vision-Language-Action Policy}

\author{Mehmet Turan Yard\i mc\i \\
Karlsruhe Institute of Technology \\
\texttt{mehmet.yardimci@student.kit.edu}
\And
Yunus Emre \c{C}o\u{g}urcu \\
\c{C}ukurova University}

\newcommand{\FactArms}{3}

\newcommand{\FactBenchmarkFamilies}{4}
\newcommand{\FactBenchmarkFamiliesWord}{four}

\newcommand{\FactGymCompleteWord}{eight}
\newcommand{\FactGymFamilies}{9}
\newcommand{\FactGymFamiliesWord}{nine}

\newcommand{\FactPairedPctEarly}{102}
\newcommand{\FactPairedPctLate}{93}
\newcommand{\FactPolicyClasses}{2}
\newcommand{\FactPolicyClassesWord}{two}
\newcommand{\FactSeeds}{3}
\newcommand{\FactSeedsWord}{three}

\begin{document}
\maketitle
\fancyhead{}
\renewcommand{\headrulewidth}{0pt}

\begin{abstract}
Online reinforcement learning (RL) fine-tuning of pretrained flow-matching vision-language-action (VLA) policies promises robots that keep learning after deployment, but continued updates frequently destroy competence on individual tasks while the aggregate still looks healthy. We study this failure mode, which we call \emph{task collapse}, under a matched small-compute budget on LIBERO-10 with a 450M-parameter SmolVLA policy trained by PPO in the stochastic (SDE) sampling regime. Three exploration-noise policies differ in a single live variable: a fixed noise scale, a learned noise network in the style of ReinFlow, and an uncertainty-gated controller that redistributes exploration across task streams from task-agnostic novelty and competence signals, without task labels or episode boundaries. Under the pooled definition, fixed noise collapses tasks in two of three seeds and learned noise in every seed measured to iteration 200, while the controller collapses none in any of its three seeds. Measured parameter displacement confirms that the controller's action expert continues to change, while its mean applied noise is close to the fixed scale in the available logs. The matched comparison supports the controller's effect on task preservation; the separate contributions of its adaptation across states and over time are not disentangled. A lower fixed scale slows the decline but does not stop it. No arm improves on the behavior-cloning baseline in this budget. Two properties of that regime are measured beside this result rather than offered as its cause: following the reference recipe, training runs in bfloat16 with no fp32 master copy, under which \num{96.02}\% of the action expert's elements stay bit-identical across three consecutive iterations, and an fp32 master copy at the reference learning rate collapses both arms in a single-seed observation. We release measurement tools for per-task collapse under four definitions, run-to-run rescoring noise and instrument tares.
\end{abstract}

\section{Introduction}
\label{sec:intro}

Robots that keep learning after deployment must preserve acquired skills while adapting to new experience. We therefore study whether adaptive exploration noise can suppress task collapse during simultaneous multi-task online RL; the longer-term fleet-learning vision is in Appendix~\ref{app:outlook}.

Two facts shaped this paper. The first is how common collapse is. Online RL fine-tuning of a pretrained policy can fall rather than plateau even under ideal conditions, a single embodiment, a fixed task set, a laboratory simulator; and if the whole policy can fall, individual tasks can fall while the aggregate still looks healthy. The second is that a task lost completely is lost in a way a task lost in part is not. A policy that still succeeds sometimes still sees reward for that task and can recover; a policy at zero may never see the reward for that task again, and rebuilding the task by imitation on top of RL can contaminate the policy it is meant to repair. Not collapsing is therefore a precondition for lifelong learning, and it was the aim of this work from the start. In our view a mechanism that keeps a policy from destroying its own competence has to exist first, before the engineering that raises the aggregate success of a policy on a task it already performs.

Our mechanism starts from a property PPO has and a flow policy lacks. A Gaussian policy exposes an explicit exploration scale: it learns an action and the width of the distribution around it, and that width varies with state. A flow-matching policy carries no such quantity, and the Flow-SDE recipe replaces it with one noise scale applied everywhere. We use novelty and value statistics to regulate the injected scale of a flow policy, from signals that read only the robot's observations and its own value estimates; we call the result uncertainty-gated exploration noise. These statistics are not, and are not claimed to be, a calibrated confidence estimate; they regulate the exploration scale.

Flow-matching and diffusion action heads are the standard parameterization of VLA policies \citep{pi0,smolvla}. Recent work has shown how to fine-tune them online with policy gradients by injecting exploration noise into the denoising process \citep{reinflow,pirl,rlinf}. The reported results are aggregate success rates. In the works we surveyed we did not find per-task dynamics during training, nor a report of how those dynamics depend on the exploration-noise policy.

We ask a narrow, matched-budget question: holding the policy, the benchmark, the PPO recipe, the data budget, and the numerical regime fixed, and changing only the exploration-noise policy, which arms keep their tasks and which collapse? We compare (i) a fixed noise scale, (ii) a ReinFlow-style learned noise network, and (iii) an uncertainty-gated controller that redistributes a bounded noise budget across task streams using two task-agnostic signals, novelty (random network distillation, RND) and competence (the PPO critic's value and the spread of a critic ensemble).

\textbf{Two contributions.}

\textbf{A. A mechanism that suppresses task collapse.} A matched-budget three-arm comparison of exploration-noise policies for online RL fine-tuning of a flow VLA, whose arms differ in one live variable and the uncertainty-gated controller is the only arm that keeps every task above its collapse threshold in every seed under the pooled definition, where fixed noise collapsed tasks in two of three seeds and learned noise in every seed. It does so without task labels or episode boundaries, and with seed-to-seed spread within the rescoring floor; aggregate success did not rise for any arm in this budget. Three seeds per arm is the design; all nine runs reached iteration 200. Supporting it: four pre-registered task-level collapse definitions, a time-to-first-collapse survival analysis, and a measurement methodology with per-instrument tares, repeated rescoring and two statistical rulers.

\textbf{B. Update absorption measured in a VLA reference recipe.} Under the reference recipe, with in-place bfloat16 parameters and no fp32 master copy, \num{96.02}\% of the action expert's elements remain bit-identical across three consecutive iterations, so most of the surface the optimizer is told to train does not move. Small-update loss under bfloat16 rounding is established for deep-network training \citep{bf16revisit}, and precision effects have been measured for language-model RL \citep{fp16mismatch,pulse}. We add the measurement for a VLA policy under the pinned reference RL recipe, together with an fp32-master pilot at the reference learning rate, reported as an observation. The pilot does not establish a suitable fp32 learning rate. This is a property of the recipe, not of our implementation, and it bounds what any arm in this regime could have shown.

\section{Related Work}
\label{sec:related}
\textbf{Online RL fine-tuning of flow and diffusion VLAs.} Flow-matching and diffusion action heads \citep{pi0,smolvla} are fine-tuned with policy gradients by making the denoising process stochastic. ReinFlow \citep{reinflow} makes the injected noise learnable and trains it jointly with the policy, which our learned-noise arm follows; $\pi_{\mathrm{RL}}$ \citep{pirl} contributes the Flow-SDE formulation our fixed and gated arms use, DPPO \citep{dppo} established the two-layer MDP view for diffusion policies, and RLinf \citep{rlinf} is the reference implementation we port. Outside robotics, Flow-GRPO \citep{flowgrpo} introduced the ODE-to-SDE conversion that makes a flow model samplable for policy gradients; FPO \citep{fpo} removes the explicit policy ratio and fine-tunes $\pi_0$ online on our benchmark, and a world-model variant reinforces flow action heads with stepwise reweighting \citep{prophrl}. A parallel line fine-tunes autoregressive VLAs from binary outcome rewards \citep{simplevlarl,riptvla}, one of them adding a learned process reward to the sparse one \citep{vlarl}, and standardizes the systems around that line \citep{rlinfvla}. All of them report aggregate success; this paper adds per-task dynamics during training and their dependence on the exploration-noise policy (Appendix~\ref{app:related}).

\textbf{Collapse and forgetting under RL fine-tuning.} Early performance drops after switching from offline to online updates are documented \citep{wsrl}, and per-task interference in multi-task learning is a classical negative-transfer phenomenon: joint training is reported to yield worse overall performance because task objectives compete \citep{taskgrouping}, and conflicting task gradients are identified as a cause of detrimental interference in multi-task supervised learning and RL alike \citep{pcgrad}. Gradient surgery acts on the gradient and our controller on exploration, so the two are orthogonal axes rather than alternatives. Two VLA methods keep an imitation term inside the online loop and state it as a stability measure, one by alternating supervised and RL phases \citep{irevla}, the other by combining behaviour cloning with Q-learning before a consistency-policy online stage \citep{conrft}.

\textbf{Exploration noise, entropy, and adaptive noise.} Maximum-entropy RL \citep{sac} regularizes stochasticity through the objective; DIME \citep{dime} and adjoint-matching methods \citep{adjointorig,adjointmatching} extend this to diffusion policies where entropy is intractable. The three lines of work above are treated at length, with the reference budgets, the matched learning rates, the absent seed statistics and the monitor and action-selector line, in Appendix~\ref{app:related}. A separate line reads a policy's own internals to predict failure at runtime, from VLA features \citep{safe} or from random network distillation combined with an action-chunk entropy score \citep{fiper}; those are the signal families used here, driving an alarm at deployment there and the exploration noise during the update here, and Appendix~\ref{app:related} states the difference in full.

\textbf{Numerical precision.} Mixed-precision training keeps an fp32 master copy so that small updates are not rounded away \citep{mixedprecision}, and nearest rounding of bfloat16 weight updates is reported to cancel small updates outright \citep{bf16revisit}; the reference VLA RL recipes we ported train the policy in bfloat16 in place and carry no precision row in their hyperparameter tables \citep{rlinf,openpi}, with the configuration evidence in Appendix~\ref{app:precision}. Precision effects have also been measured in language-model RL \citep{fp16mismatch}. \citet{pulse} report that about \num{99}\% of per-step weight updates are invisible after the bfloat16 cast and attribute it to Adam updates falling below the local rounding threshold; that setting keeps fp32 master weights and casts them for the forward pass, so it is a different regime from the in-place update without a master copy measured here, and the quantity it reports is the one this paper states it did not measure. We found no equivalent measurement for a robot policy and supply one in Section~\ref{sec:analysis}.

\section{Method}
\label{sec:method}
\subsection{Flow-matching policy and SDE sampling}
The policy $\pi_\theta$ maps an observation $o_t$ (two camera images, proprioception, language instruction) to an action chunk by integrating a learned velocity field from Gaussian noise $x_0$ to $x_1$ over $K=10$ denoising steps. Of the integrated chunk, 50 steps of a 32-dimensional padded action, the first $H=5$ steps over the $d=7$ controlled dimensions are executed open loop and form the action $a_t \in \mathbb{R}^{H \times d}$ whose log-density enters the policy-gradient ratio. Deterministic (ODE) integration yields a single action; following the Flow-SDE construction \citep{pirl,dppo} we add Gaussian noise of scale $\sigma$ at each denoising step so that each step is a Gaussian transition with a tractable log-density, and the full trajectory becomes a two-layer MDP over environment steps and denoising steps. All evaluations in this paper use the stochastic sampler with $\sigma_{\mathrm{eval}}=0.25$; the deterministic sampler is never used as a decision ruler, because the policy this work builds toward stays under online RL and therefore never runs in ODE mode.

\subsection{PPO on the two-layer MDP}
We use PPO \citep{ppo} with the reference operators \citep{rlinf}: clipped surrogate with dual clipping, value clipping, Huber value loss, truncation compensation and gradient clipping. PPO trains one value head and computes advantages from it alone; the controller's five-head ensemble (Section~\ref{sec:controller}) enters neither the advantage nor the PPO loss (Appendix~\ref{app:controller}). The algorithm comes with the recipe and was not selected here, and an empirical study of VLA RL fine-tuning reports PPO as more effective for this model class than language-model-derived alternatives \citep{rlvlagen}.

\subsection{Exploration-noise arms}
The three arms differ in one live variable, the SDE noise scale $\sigma$ used during data collection:
\begin{itemize}
\item \textbf{F (fixed):} $\sigma = 0.25$ for all tasks and all iterations. Ablations use $\sigma \in \{0.18, 0.15\}$.
\item \textbf{R (learned):} a noise network in the style of ReinFlow \citep{reinflow} predicts $\sigma$ per denoising step and is trained jointly. Read from the module source, it is a two-layer MLP of width 64 with $\tanh$ activations over the concatenation of a learned 16-dimensional denoising-step embedding and the pooled VLM prefix embedding, whose log-variance output is $\tanh$-squashed into the bounds $[0.10, 0.24]$ taken from the reference configuration. It holds \num{170848} trainable parameters, \num{0.038}\% of the policy's published 450M size, counted by instantiating the module on CPU at the geometry the runs used.
\item \textbf{K (controller):} a per-stream noise scale taking one of the four values $\{0.12, 0.30, 0.36, 0.42\}$, recomputed at every environment step from task-agnostic signals and applied at the following step.
\end{itemize}

\subsection{The controller}
\label{sec:controller}
\textbf{Why these signals, and why not success rate.} The controller exists to produce the quantity a flow policy does not carry: how sure the policy is, at this state, about which action is good. The reasoning behind the four cells, the real-world constraints behind them, the eight-iteration hold-off on the competence signal and the code-level proof that no task label is read are in Appendix~\ref{app:controller}.

The controller holds one noise scale per rollout stream and recomputes it at every environment step
from that step's features. It carries state while doing so: the statistics that standardize those
features are exponential moving averages with decay \num{0.99}, pooled over slots rather than kept
per slot, and the novelty predictor is trained once per outer iteration, so the same feature vector
does not score the same twice.
 Write $i$ for a slot of the vectorized
environment, $z_i$ for the whitened random-network-distillation score of the last denoising step's
feature vector \citep{rnd}, $v_i$ for the standardized value of the PPO critic, $s_i$
for the standardized spread of a separate five-head value ensemble that the controller trains, $m$ for the number of observations the novelty statistic
has seen, and $o$ for the outer iteration. Two binary signals are formed,
\begin{align}
n_i &= \mathbf{1}[z_i > 1.28] \wedge \mathbf{1}[m \ge 512], \label{eq:novel}\\
\ell_i &= \left(\mathbf{1}[v_i < -0.5] \vee \mathbf{1}[s_i > 1.0]\right) \wedge \mathbf{1}[o \ge 8], \label{eq:lowcomp}
\end{align}
and they index a $2\times2$ multiplier table $M$ applied to the base scale $\sigma_0 = 0.3$,
\begin{equation}
\sigma_i^{(t+1)} = \mathrm{clip}\!\left(\sigma_0\, M[n_i, \ell_i],\; 0.1,\; 0.7\right),
\label{eq:sigma}
\end{equation}
with $M[\text{novel},\text{low}] = 1.4$, $M[\text{novel},\text{high}] = 1.0$,
$M[\text{familiar},\text{low}] = 1.2$ and $M[\text{familiar},\text{high}] = 0.4$. The four
reachable values are therefore $\{0.12, 0.30, 0.36, 0.42\}$, so the band is
$[0.12, 0.42]$ and the clip in \eqref{eq:sigma} alters none of them. The superscript in
\eqref{eq:sigma} is a lag: the scale computed from step $t$'s features is applied at step $t+1$,
and the first step of each rollout buffer is sampled at $\sigma_0$.

\section{Experimental Setup}
\label{sec:setup}
\textbf{Policy and benchmark.} SmolVLA (450M) \citep{smolvla} fine-tuned by behavior cloning on LIBERO-10 \citep{libero}, the long-horizon suite. Pool tasks t0--t6 are \texttt{libero\_10:\{1,2,3,5,6,7,9\}}, in that order.

\textbf{Why this suite and not four.} The four LIBERO suites are not equally informative for this question. The behavior-cloning weights every arm starts from score \num{0.5762} on LIBERO-10 and between \num{0.8333} and \num{0.9444} on the other three, so on those a task starts near the ceiling, with little room above it and its collapse threshold, half its own tare, far below. This study asks both whether tasks erode and whether they are held, so it needs room in both directions, and LIBERO-10 is the suite that has it. The per-suite tares, the shorter ruler that keeps those columns from being subtracted from ours, and the same rule applied again inside LIBERO-10 to select the seven-task pool are in Appendix~\ref{app:faq}.

\textbf{Budget.} 200 outer iterations per arm, 2044 samples per iteration, four optimizer steps per iteration.

\textbf{Two rulers.} A single 210-episode scan carries a Wilson 95\% interval of half-width about 0.067. Rescoring the same weights, changing only the evaluation seed, gives a run-to-run standard deviation of \num{0.0375} on the campaign instrument (four scans of the controller seed-3337 iteration-200 checkpoint: 0.6048, 0.6286, 0.5619, 0.5476). A second four-scan set on the same instrument gives \num{0.0147}, so this floor depends on the checkpoint as well; we carry the larger throughout, and the local instrument gives \num{0.0198} over three scans of the BC weights. A pooled interval over 1050 episodes rests on an assumption: Wilson treats those episodes as independent Bernoulli trials of one rate, while they are five scans of seven tasks whose rates differ and which are ordered in training time, so it bounds the sampling of the pooled quantity and not the variation between tasks or across the window. Both rulers are given for every comparison, because the interval describes one scoring and the run-to-run SD describes repeatability, and neither contains the other: on the local instrument the SD reaches \FactPairedPctLate{} and \FactPairedPctEarly{} percent of one scan's Wilson half-width at the two checkpoints where it was measured, so the two are reported separately. The two instruments' tares, the episode-identity measurement across evaluation geometries and the paired rescoring study are in Appendix~\ref{app:repro}.

\textbf{Collapse metrics (pre-registered).} A task is \emph{collapsed} when its success rate falls below one half of the same task's tare on the same instrument. We report four definitions: pooled collapse over the last five canonical scans (150 episodes per task), last-scan collapse (30 episodes per task), sustained collapse over two consecutive scans, and ever crossing the threshold at any measured scan. Pooled collapse is primary; the survival analysis uses the time of the first single-scan crossing. Time to first collapse is analyzed as a survival curve with right censoring at the last measured scan.

 Success rate is not the driver, by design: the controller spreads exploration across tasks rather than concentrating it where reward is already plentiful. The reasoning is in Appendix~\ref{app:controller}.

\section{Results}
\label{sec:results}
\subsection{Learning curves}
\begin{figure}[!b]
\centering
\includegraphics[width=0.85\columnwidth]{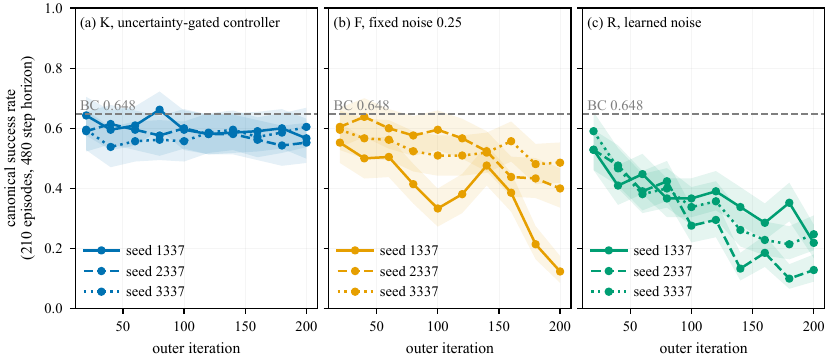}
\caption{Canonical success rate (210 episodes, horizon 480, SDE sampling with $\sigma_{\mathrm{eval}}=0.25$) against outer iteration, one line per arm and seed; dotted lines are the tare of each instrument. Two rulers: the shaded band is the Wilson 95\% interval of a single 210-episode scan (half-width about 0.067), and the bar at the right margin is the run-to-run SD of rescoring the same weights, 0.0375 on the campaign instrument over four scans and 0.0198 on the local instrument over three. The second ruler is not contained in the first; a gap smaller than both is not read as a difference.}
\label{fig:curves}
\end{figure}

\begin{figure}[t]
\centering
\includegraphics[width=0.85\columnwidth]{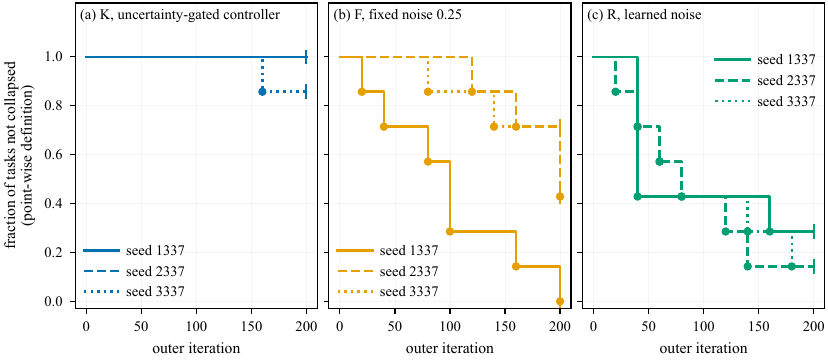}
\caption{Fraction of the seven tasks not collapsed by each outer iteration under the point-wise definition (a task is collapsed when one 30-episode scan falls below half of its own tare on the same instrument), one curve per arm and seed, right-censored at each run's last measured scan. Two rulers: the point-wise definition is read from 30 episodes per task, Wilson 95\% half-width about 0.17, and repeat scoring moves the pooled rate by a run-to-run SD of 0.0375 on the campaign instrument. The figure is for the shape over many iterations; an isolated crossing is read cautiously.}
\label{fig:survival}
\end{figure}

Fig.~\ref{fig:curves} shows canonical success against iteration for all arms and seeds with each arm's own tare. The three fixed seeds diverge from one another within the first hundred iterations and two of them end far below the tare; all three learned-noise seeds fall early, each reaching its first collapse within the first forty iterations, though their curves after that point are not identical; the three controller seeds stay within a band of the tare for the whole run and within the larger of the two measured rescoring floors of one another. Time to first collapse is shown in Fig.~\ref{fig:survival}.

\subsection{Plateau pools}
Table~\ref{tab:pool} gives each run's pooled success, Wilson interval and four collapsed-task counts.

\begin{table}[t]
\caption{Plateau pool over the last five canonical scans, each arm against its own instrument tare (campaign tare 136/210 $=$ 0.648, local tare 121/210 $=$ 0.576). Four collapse columns share one threshold, a task below half of its own tare, and differ where a task crossed and later recovered: \emph{Pool} is the primary definition, \emph{Point} the single-scan one the survival analysis uses, \emph{Sust.} two consecutive scans, \emph{Ever} any scan. All four are stated formally in Appendix~\ref{app:definitions}. \emph{Last} is each run's last measured scan; a run censored early has had fewer opportunities to collapse. Every campaign row is scanned every \num{20} iterations and pools iterations \num{120} to \num{200}. The one exception is the local-instrument row in the last block, whose iteration-160 scan is a \num{14}-episode fragment the \num{210}-episode rule excludes, so it pools \num{100} to \num{200}; that row is marked not comparable to the others. The fp32 rows are a pilot and belong with the analysis. Every row is a single seed and carries no arm-level verdict.}
\label{tab:pool}
\centering
\small
\setlength{\tabcolsep}{4pt}
\resizebox{\textwidth}{!}{%
\begin{tabular}{llrrlcccc}
\toprule
Arm & Seed & Last & Pool SR & Wilson 95\% & Pool & Point & Sust. & Ever \\
\midrule
K controller & 1337 & 200 & \num{0.585} & \num{[0.555, 0.614]} & \num{0/7} & \num{0/7} & \num{0/7} & \num{0/7} \\
K controller & 2337 & 200 & \num{0.564} & \num{[0.534, 0.594]} & \num{0/7} & \num{0/7} & \num{0/7} & \num{0/7} \\
K controller & 3337 & 200 & \num{0.589} & \num{[0.559, 0.618]} & \num{0/7} & \num{1/7} & \num{0/7} & \num{1/7} \\
F fixed 0.25 & 1337 & 200 & \num{0.316} & \num{[0.289, 0.345]} & \num{4/7} & \num{7/7} & \num{6/7} & \num{7/7} \\
F fixed 0.25 & 2337 & 200 & \num{0.472} & \num{[0.442, 0.503]} & \num{1/7} & \num{4/7} & \num{2/7} & \num{4/7} \\
F fixed 0.25 & 3337 & 200 & \num{0.510} & \num{[0.480, 0.541]} & \num{0/7} & \num{0/7} & \num{1/7} & \num{2/7} \\
R learned    & 1337 & 200 & \num{0.317} & \num{[0.290, 0.346]} & \num{4/7} & \num{5/7} & \num{4/7} & \num{5/7} \\
R learned    & 2337 & 200 & \num{0.169} & \num{[0.147, 0.192]} & \num{6/7} & \num{6/7} & \num{6/7} & \num{6/7} \\
R learned    & 3337 & 200 & \num{0.262} & \num{[0.236, 0.289]} & \num{5/7} & \num{5/7} & \num{5/7} & \num{6/7} \\
\midrule
\multicolumn{9}{l}{\emph{Noise-scale ablation, single-seed observations}} \\
F fixed 0.18 & 1337 & 200 & \num{0.496} & \num{[0.466, 0.526]} & \num{1/7} & \num{2/7} & \num{1/7} & \num{3/7} \\
F fixed 0.15 & 1337 & 200 & \num{0.559} & \num{[0.529, 0.589]} & \num{0/7} & \num{0/7} & \num{1/7} & \num{1/7} \\
F fixed 0.15 & 2337 & 200 & \num{0.463} & \num{[0.433, 0.493]} & \num{2/7} & \num{3/7} & \num{2/7} & \num{4/7} \\
\midrule
\multicolumn{9}{l}{\emph{fp32 master-copy pilot, single seed each, both measured to iteration 200}} \\
K controller & 1337 & 200 & \num{0.283} & \num{[0.256, 0.311]} & \num{3/7} & \num{5/7} & \num{3/7} & \num{7/7} \\
F fixed 0.25 & 1337 & 200 & \num{0.086} & \num{[0.070, 0.104]} & \num{7/7} & \num{7/7} & \num{7/7} & \num{7/7} \\
\midrule
\multicolumn{9}{l}{\emph{Local instrument, not comparable to the rows above}} \\
K controller & 1337 & 200 & \num{0.514} & \num{[0.484, 0.544]} & \num{1/7} & \num{3/7} & \num{2/7} & \num{3/7} \\
\bottomrule
\end{tabular}
}
\end{table}

\subsection{Task collapse and survival}
\label{sec:collapse}
The four definitions distinguish persistent losses from transient threshold crossings. Counting any scan, every seed of the fixed and learned arms has crossed the threshold on at least one task, and so has one of the three controller seeds. Pooled over the last five canonical scans, two of three fixed seeds and all three learned seeds have at least one collapsed task, while fixed seed 3337 and the $\sigma=0.15$ ablation at seed 1337 have none, because the tasks they lost earlier recovered enough to lift the pooled window back above the threshold. The gap between the definitions is not small: fixed seed 1337 stands at 4 of 7 pooled against 7 of 7 on its last scan, and fixed seed 2337 at 1 against 4. A single scan uses 30 episodes per task against 150 in the pooled window; its threshold crossings are reported beside the pooled, sustained and ever counts. First-collapse iterations carry what the seed-level test cannot: 20, 120 and 80 for the three fixed seeds, 40, 20 and 40 for the three learned seeds, and 160 for the one controller seed that crossed, with the other two censored at 200 without an event. The fourth definition, sustained collapse over two consecutive scans, sits between the pooled and the ever counts and provides complementary evidence: every fixed and learned seed contains a sustained collapse (3/3 and 3/3, and 6 of 7 tasks on fixed seed 1337), whereas none of the three controller seeds does (0/3). The pooled definition remains the primary one and the sustained count is reported beside it. At the seed level a Fisher exact test on the pooled definition gives $p = 0.400$ for fixed noise against the controller (2/3 against 0/3); with three seeds per arm its smallest attainable value is $p = 0.100$, at 3/3 against 0/3, so no split reaches 0.05. The other two seed-level tests, the two cases that show why all four definitions are reported rather than one, and the scan-cadence caveat on the sustained column are in Appendix~\ref{app:definitions}.

\subsection{Seed variance}
At $\sigma=0.25$ the pooled plateau spans \num{0.3162}, \num{0.4724} and \num{0.5105} across the three fixed seeds, a spread of \num{0.194} taken at that precision rather than from the rounded figures, more than five times the run-to-run rescoring SD, so it is a property of the seeds and not of the instrument. The three controller seeds pool at \num{0.585}, \num{0.589} and \num{0.564}, a standard deviation of \num{0.013} with the widest pair \num{0.0248} apart. That spread is inside the \num{0.0375} floor and just outside the \num{0.0147} one, so we claim only what survives both, the contrast with the fixed arm, whose seed spread of \num{0.194} is more than ten times the smaller floor; per-seed curves are in Appendix~\ref{app:pertask}.
\section{Analysis: is the protection just less exploration?}
\label{sec:analysis}
\textbf{Mean noise.} The controller's mean applied noise is \num{0.2399} over the local run's 144 logged outer iterations, \num{0.2444} over all 200 of campaign seed 1337 and \num{0.2353} over the 155 logged for seed 2337, against the fixed arm's \num{0.2500} at a standard deviation of \num{0.0000}; the learned arm reads \num{0.2340} over 146 outers that begin at iteration 54, so its early iterations are absent and its range is not matched to the others. The gap to the fixed scale is \num{0.006} to \num{0.015} for the controller by run and coverage and \num{0.016} for the learned arm, reported beside the result rather than as a precondition for it.

\textbf{Noise allocation across tasks.} Campaign logs also show differentiated noise allocation across tasks. The three tasks with lower baseline success receive mean scales of \num{0.288} and \num{0.274} in seeds 1337 and 2337, against \num{0.180} and \num{0.185} for the three with higher baseline success. Coverage is 200 of 200 outer iterations for seed 1337 and 155 of 200 for seed 2337; the log for seed 3337 was unavailable. The task label enters only this post-hoc grouping and is not an input to the controller, and the logged field is aggregated per task and per outer iteration, so nothing is claimed about behaviour within an iteration. Grouping, weighting and the missing iterations are in Appendix~\ref{app:pertask}.

\textbf{Lower fixed noise.} At seed 1337, fixed scales of 0.18 and 0.15 pool at \num{0.496} and \num{0.559}, with one and zero collapsed tasks respectively, and both exceed the fixed-0.25 result on that seed. The second fixed-0.15 seed pools at \num{0.463} with two collapsed tasks, so the first seed's preservation does not replicate. The 0.18 ablation has one seed and 0.15 has two. No figure plots them; see Table~\ref{tab:pool}, Appendix~\ref{app:faq} for the first 0.15 seed's last three readings, and the archive's \texttt{evaluation/canon}.

\textbf{Precision.} Following the reference recipe, trainable parameters are stored and updated in bfloat16 with no fp32 master copy \citep{rlinf,openpi}, a departure from the mixed-precision convention of keeping an fp32 master \citep{mixedprecision}; the exact configuration lines are in Appendix~\ref{app:repro}. Measured across three consecutive iterations, \num{96.02}\% of the \emph{elements} of the action expert body, its individual weight entries, are bit-identical, and the frozen side is bit-identical in \num{100}\% of its elements, which is the null control that shows the comparison itself works. Over the same three iterations \num{112} of the expert's \num{145} \emph{tensors} change in at least one element. The fraction of parameters whose Adam step falls below bfloat16 resolution is a different quantity and was not measured. A second percentage counts whole tensors rather than elements over 40-iteration windows: across the \num{33} windows that could be opened, \num{11.9} to \num{13.3}\% of the trainable tensors are bit-identical end to end under bfloat16, against \num{1.4}\% in the fp32-master pilot. Over the same windows the action expert moves \num{0.0008} to \num{0.0010} under bfloat16 against \num{0.0032} to \num{0.0035} with the master copy. Across the measured windows, all three arms show action expert displacement in the same narrow range. Task preservation in the controller arm therefore occurs while its action expert continues to change (Appendix~\ref{app:movement}). Appendix~\ref{app:precision} reports the fp32-master pilot and the limits of that comparison.

\section{Discussion}
\label{sec:discussion}
\textbf{Reading the results.} The controller's advantage is not uniform superiority but protection where the baselines break. In the fixed-noise arm the seed that lost six of its seven tasks raised the seventh above its tare while doing so. This pattern is consistent with interference between tasks, but gradient competition was not measured directly; the matched intervention evaluates the exploration controller as a whole. The fixed-scale results vary across seeds, whereas the controller endpoints lie within the larger measured rescoring SD of one another. Under the fp32 pilot, where the same rate bites much harder, the controller kept more tasks than the fixed arm at equal exposure (3/7 against 7/7 collapsed), an observation in a pilot with one seed per arm.

\textbf{Why no gain here.} This budget did not raise aggregate success: fourteen environments feed a sparse success signal, a 450M-parameter policy stores its competence compactly and is fragile to small displacements, LIBERO-10 is the hardest suite, and the reference learning rate appears stable in bfloat16, where \num{96.02}\% of the action expert's elements stay bit-identical across three consecutive iterations (Section~\ref{sec:analysis}), read against the one-seed fp32 pilot (Appendix~\ref{app:precision}). None of the four is offered as the explanation; this design cannot separate them. The fp32-master pilot's collapse at the reference rate motivates retuning it; an appropriate fp32 rate was not established here.

\textbf{Breadth, and what it scopes.} The reference mechanism was re-run on its own benchmarks at matched budget: \FactGymCompleteWord{} complete tasks over \FactPolicyClassesWord{} policy classes with three seeds per arm, nine attempted. The screening rule asks an arm's mean to lie outside the other arms' seeds and its difference to exceed both the seed spread and the largest within-evaluation spread. Learned against fixed, it fires on no task, a limit of resolution and not an equivalence (on Can, \num{+4.60} clears \num{4.10}, the largest seed SD among the three arms, not the within-evaluation spread of \num{41}). Controller against fixed, it fires on three as the owner record applies it (Hopper and Kitchen mixed down, Kitchen complete up); a stricter reading, applied to every task, does not keep the upward one, so two remain, both downward, and no positive separation survives (Appendix~\ref{app:reinflowgym}). These settings do not reproduce the simultaneous language-conditioned mixture of LIBERO, so a benefit from reallocating noise across task streams need not follow here; a hypothesis for the Kitchen mixed drop, with stage-level behaviour not analyzed, is in Appendix~\ref{app:reinflowgym}.

\textbf{Why simultaneous rather than sequential.} If the experience of a fleet arrives at once, many tasks flow into the same gradient as a continuing mixture, which is not the regime the sequential forgetting experiment describes. This paper is not a lifelong-learning result, and its claim is the narrower one that Section~\ref{sec:limitations} states first: exploration, update size and plasticity stand together in that program and this paper measures the first. The fleet-level design is in Appendix~\ref{app:outlook}.

\section{Limitations}
\label{sec:limitations}
Each item names a limit and what it rests on, so it can be checked at its source.
\begin{itemize}
\item \textbf{Necessary, not sufficient.} Resistance to per-task collapse under simultaneous multi-task training is a necessary property for a policy that keeps learning without losing what it does; collapse is the quantity we measure and its absence is what the property asks for. The sequential lifelong setting is not measured here (Appendix~\ref{app:repro}).
\item \textbf{No gain over behavior cloning.} No arm exceeds its own tare. Raising aggregate success was not the objective and no arm was tuned for it; the mechanism was pre-registered to reduce per-task collapse, which is what the three controller seeds measure, on one policy, one suite and one budget. The tare is near the LIBERO-10 level published for SmolVLA at this size \citep{smolvla} and below it on both instruments with the Wilson intervals overlapping, so the room above it is small by construction; a sparse success signal, 800 optimizer steps and bfloat16 in-place updates stand with it. The published figure and both gaps are in Appendix~\ref{app:repro}.
\item \textbf{Precision and learning rate.} Training ran in bfloat16 with no fp32 master copy as in the reference recipes; we measured the consequence (96.02\% of action-expert elements bit-identical over three iterations), and the fp32 pilot holds the reference learning rate fixed, so the right fp32 rate is not established (Appendix~\ref{app:precision}).
\item \textbf{Breadth.} Task preservation is shown for one VLA policy on seven jointly trained LIBERO tasks. Across the \FactGymCompleteWord{} complete tasks of Appendix~\ref{app:reinflowgym} the controller shows no positive separation that the stricter reading retains, and two negative ones. Generalization to other VLA policies and to other training regimes is untested.
\item \textbf{The learned-noise family is represented by one member.} The Flow-Noise variant of \citep{pirl} was not run; our R arm is one member of that family, not its best case.
\item \textbf{Evaluation budget and instrument.} A single canonical scan uses 30 episodes per task, below the de facto LIBERO standard of 50, though the pooled definition that carries the main result averages the last five scans and rests on 150; the same weights score 15 episodes apart on two GPUs, so every comparison is read within one instrument (Appendix~\ref{app:repro}).
\item \textbf{Scope of the evaluated controller.} The reported effect is obtained with two signal families and a four-cell rule. Additional drivers present in the code were inactive, so the results apply to the evaluated rule. One cell of the multiplier table departs from the design record and is named in Appendix~\ref{app:controller}. The inactive drivers are untested here rather than shown ineffective.
\item \textbf{Exposure.} Every arm of the LIBERO campaign was measured to iteration 200 and scanned every \num{20} iterations, so no campaign run enters a seed-level count with fewer opportunities to collapse than another; the local-instrument row of Table~\ref{tab:pool} and the re-run of Appendix~\ref{app:reinflowgym} sit outside that scope and are marked where they appear. How one controller seed reached 200, and what that leaves unmeasured, is in Appendix~\ref{app:repro}. The seed-level counts therefore compare three complete fixed seeds against three complete controller seeds, and both tables that report run outcomes carry each run's last scan.

\end{itemize}

\section{Conclusion}
\label{sec:conclusion}
We asked what happens to individual tasks when a pretrained flow-matching VLA is fine-tuned online under a small matched budget, and how that depends on the exploration-noise policy. Task collapse is the dominant phenomenon: fixed noise loses tasks at a seed-dependent rate, and the evaluated learned-noise arm early and systematically. Within the matched LIBERO training regime, adaptive exploration noise suppressed task collapse: no controller seed exhibited pooled collapse, compared with two of three fixed seeds and all three learned seeds. No controller seed exhibited sustained collapse either. These results support task preservation during continued updates in the evaluated regime, without an aggregate gain over behavior cloning.

\section*{Reproducibility Statement}
Every number in this paper is read from a measurement file named in the text or in Appendix~\ref{app:repro}, which lists the run manifests (trainable surface, batch arithmetic read from the run's own counter, precision of parameters and optimizer state), the evaluation protocol (210 fixed-identity episodes per canonical scan, SDE sampling with $\sigma_{\mathrm{eval}}=0.25$, horizon 480), the two rulers (per-scoring Wilson interval and repeat-scoring SD, measured on each instrument), the four collapse definitions and their thresholds, the seven-task pool and the reason for the three exclusions, and the pinned reference configuration from which the PPO operators and learning rates were matched. The code and the measurement files are submitted as anonymized supplementary material: the canonical evaluation tree and every repeat scan, the measurement and gate scripts named throughout this paper, the controller module and the training code the campaign ran, the campaign configuration, the run configurations of the seed-1337 controller pair with their measured batch arithmetic, the launch banner and trainable-surface manifest of the 14-environment local run, two preflight records from before the 14-environment geometry, named as such, which fix the trainable surface and not the batch, and the owning measurement files from which every number here is read. Table~\ref{tab:supplementary} names every measurement file the archive carries and how each was made. The model weights are too large for the archive and are released at an anonymous repository instead: every checkpoint the runs wrote that a copy still holds, for every arm reported here but the one named below, together with the behavior-cloning weights each run started from. Every canonical success rate in this paper is read at an outer iteration that is a multiple of twenty, which is where a canonical evaluation happens; the checkpoints between them were scored by no evaluation and are released all the same. One reading comes from outside the released arms: Tables~\ref{tab:round} and~\ref{tab:refcfg} read iterations 33 and 36 of \texttt{k\_corrected\_s1337}, an earlier seed-1337 controller run on the same seven tasks and from the same behavior-cloning weights (equal sha256), with one environment per task, four accumulation rounds and the same 2044 samples per optimizer step, which stopped at iteration 38. Those two checkpoints are in the anonymous repository; the run configuration its resumed process wrote is in the archive as \texttt{run\_config\_k\_corrected\_s1337.json}, with an empty measured block because that process trained no step, and the run's measured batch, its checkpoint lines and its resumes are in \texttt{manifests/measured\_k\_corrected\_s1337.log}, lines taken from the run's own log. The archive carries their manifest, which gives the address, and the name, arm and outer iteration of every released checkpoint, and names the checkpoints that exist in no copy together with the reason each is absent; the sha256 of each file is shown beside the file at the repository. It also carries the script that wrote the files, the command that checks a download against the manifest, and the command that turns a downloaded file back into a checkpoint file holding the model state alone, which the trainer loads for scoring with its \texttt{--eval\_ckpt} option; training does not resume from a released file, because a resume also reads the optimizer and controller state, which the release does not carry.

\section*{Statement on the Use of AI}
We used generative AI tools (Claude, Anthropic; Codex, OpenAI) to assist with implementing methods and writing and editing software code (the training infrastructure, the controller and the measurement scripts, under the authors' specification), with literature search and summarization, with drafting parts of the paper and translating the authors' notes, with formatting references, and with producing scientific figures from measurement files. AI tools also gave feedback on hypotheses, on conceptual framing, on experimental methodology and protocol design, and on the interpretation of quantitative results; the authors made the final research and interpretation decisions. We did not use AI tools to generate synthetic data, to formulate mathematical claims, to supply proof ingredients or write proofs, or to clean and reformat datasets. Formal qualitative or thematic analysis of data was not part of this study, so that category does not apply; AI assistance with interpreting quantitative results is disclosed above. AI-assisted code and analyses were reviewed with automated tests and planted-violation checks. Numbers, citations and claims were checked against their owning measurement files, the code and the cited sources by scripted checks and by AI-assisted audits, and the authors reviewed the results of those checks. The authors take responsibility for the final content of this work.

\bibliography{references}
\bibliographystyle{iclr2027_conference}

\appendix
\section{Extended related work}
\label{app:related}
These works report aggregate suite-level success, and $\pi_{\mathrm{RL}}$ reports it by suite: for $\pi_0$, few-shot supervised fine-tuning gives 65.3, 64.4, 49.8 and 51.2 percent on spatial, object, goal and long, and its Flow-Noise variant gives 99.0, 99.2, 98.2 and 93.8, with the per-suite breakdown in its Appendix Table 3. Its LIBERO budget is listed in Appendix Table 11 (Appendix J): 500 training epochs, a global batch of 2048, 64 parallel environments, 8 rollout epochs, 4 update epochs, and 240, 240, 320 and 480 interaction steps across the four suites; the reported hardware is eight NVIDIA H100 80GB GPUs. The two learning rates we matched, actor $5\times10^{-6}$ and critic $10^{-4}$, are not in those tables; we read them from the pinned RLinf LIBERO configuration and mark them as a code-level secondary source rather than a claim about the paper. A number of random seeds for the LIBERO experiments is not stated anywhere in that paper, and no standard deviation, confidence interval or error bar appears in any of its results tables; the only sample size it reports is its evaluation protocol, 500 initial states per suite. A configuration posted in the RLinf issue tracker carries a single \texttt{seed: 42}, which is evidence about one configuration file and not about the paper's experiments. We take that absence as the concrete instance of a general pattern, and report per-task dynamics and seed spread because they are what our question is about.

\textbf{Monitors, action selectors, and noise scales.} A growing line of work reads a confidence or failure signal out of a VLA, and it is worth separating from ours by what the signal is allowed to do. VLAConf \citep{vlaconf} fits a calibrated task-success probability on top of frozen pretrained representations, explicitly to serve flow-matching policies that expose no action-token probabilities, and uses it on LIBERO to trigger handoffs to an expert. Tri-Info \citep{triinfo} derives three information-theoretic signals for failure detection and reports that they transfer across architectures and to real robots without retraining. Both observe a policy they do not change: the output is a probability or an alarm, and the action distribution is untouched. DF-ExpEnse \citep{dfexpense} does intervene, and closer to us, but at a different point: it builds a candidate action set from the generative policy's own multimodality and uses a critic ensemble to pick the candidate that best trades quality against exploration interest during fine-tuning. MG-Select \citep{mgselect} selects among candidates too, at test time rather than during fine-tuning, and it scores them by the KL divergence of the policy's own action-token distribution from a reference obtained by masking the state and language inputs, which is a quantity a flow-matching action head does not produce. Our mechanism selects no candidate; it moves the scale of the injected noise, which is the width of the distribution rather than a choice within it, and it is the only one of the four that reports a per-task collapse metric under continued updates.

ReinFlow \citep{reinflow} injects learnable noise into the flow steps and trains it jointly: its noise standard deviation is produced by a network conditioned on the denoising time, the current action and the observation, and its bounds are set from the mechanical limits of each joint. It prescribes no schedule over training time; the paper reports instead that the noise level decays of its own accord once the success rate saturates. The contrast with a signal-driven scale is therefore about what drives the decay, return in one case and novelty and competence in the other, and not about scheduled against unscheduled. $\pi_{\mathrm{RL}}$ \citep{pirl} proposes two variants. Flow-Noise takes each denoising step to be one step of a discrete-time MDP and learns the noise added at that step with a small network, which is what lets it compute an action log-likelihood exactly instead of approximating one. Flow-SDE instead replaces the ODE sampler with an SDE and nests the denoising steps inside the environment steps, giving a two-layer MDP. It reports large aggregate gains on LIBERO. RLinf \citep{rlinf} provides the reference implementation we port. DPPO \citep{dppo} established the two-layer MDP view for diffusion policies. Our fixed and controlled arms are Flow-SDE style; our learned arm follows ReinFlow. The reference budgets, the matched learning rates and the absent seed statistics are recorded in this appendix.

In the RL fine-tuning of \emph{reasoning language models}, a domain distinct from ours, policy entropy is observed to contract sharply early in training, and that contraction is reported as accompanying the saturation of policy performance rather than as a leading indicator of a later collapse \citep{entropymech}. We cite it as an analogy for a mechanism, not as evidence about robot policies. RL fine-tuning of VLAs is often reported without seed statistics, as in the case named above, though we can name instances rather than survey the field and do not put this forward as a universal claim; Henderson et al. document substantial variability across random seeds and do not recommend a universal number of trials \citep{deeprlmatters}.

MaxEntDP \citep{maxentdp} realizes the maximum-entropy objective with a diffusion policy on MuJoCo; its temperature coefficient is chosen per task and held fixed for the whole of training, and its own limitations section names automatic adaptation of that coefficient as future work. It is the closest published relative \emph{in purpose}: it is the work whose knob is the same knob, an explicit exploration level for a generative policy, and the difference is exactly what our arms turn on, since there the level is set per task before training and frozen, while here it is a bounded scale moved during training by signals that never read a task label. State-dependent exploration \citep{gsde} and, recently, state-aware exploration regulation for flow policies \citep{flowecer} adapt noise from policy statistics. The closest published work \emph{in intervention point} is \citep{flowecer}, whose Entropy-Covariance Exploration Regulator sets the exploration level of a flow policy from policy statistics during training, which is where our controller also acts. Read from its method, experiment and result sections rather than from its abstract, five differences from our setting stand, each of which we rely on. It is evaluated on DeepMind Control Suite, Humanoid-Bench and MuJoCo Gym, which are state-based continuous-control suites; its policy is neither vision-language nor language-conditioned; its setup describes no pretrained initialization and no demonstration data, so it is not fine-tuning a pretrained policy; its signals are policy entropy and action-uncertainty covariance, whereas ours are random-network-distillation novelty together with the critic's value and the spread of a critic ensemble, and the terms \emph{ensemble} and \emph{critic ensemble} do not occur in its text; and it reports normalized scores rather than per-task success, with neither \emph{success rate} nor any per-task collapse metric appearing in it. The two works are therefore close in different respects, one in purpose and one in intervention point, and neither is nearer than the other on a single ranking. No experiment here runs its regulator: its full text names no code release, and it specifies the regulator for state-based control without a pretrained initialization, so a port to a pretrained vision-language-action policy would be a design of ours rather than a reproduction of theirs. Of its two signals, the policy entropy is available here in principle, since it is estimated from a diagonal mixture fitted to actions sampled at one state and a flow policy can be sampled repeatedly at one state, though we did not compute that estimate; the action-uncertainty covariance is missing, since it correlates that mixture's log-density with the spread of return samples from a distributional critic conditioned on state and action, and both critics here, the PPO critic and the five-head ensemble, are state-value heads with no per-action return spread. Confidence signals have been used as an intrinsic reward for VLAs \citep{t2vla}, where the confidence-success correlation is reported for discrete-action VLAs; we use confidence-type signals as a noise scale, not as a reward. The monitor and action-selector line of work that reads confidence out of a VLA without changing its action distribution is separated from ours in this appendix.

\citet{pistepnft} is the most recent online RL method for flow-based VLAs and is built on the same reference implementation we port; it sweeps the SDE noise level as a fixed hyperparameter, states no number of random seeds, and reports no per-task quantity, which is the measured form of the gap this paper's introduction names. \citet{robustvla} scales injected observation and action noise by the policy's smoothed success rate as a robustness curriculum; it is a published VLA noise schedule driven by exactly the signal Appendix~\ref{app:controller} argues against for exploration, and the difference is that it is a single global level raised or lowered over training rather than a per-stream allocation, and that its purpose is perturbation robustness rather than where exploration is spent. Success rate is a task-level average that is known only when an episode ends, so it lags the policy it describes, while the competence signal is read at every environment step and is specific to the current state. A driver built on success rate therefore piles exploration onto the task whose reward is plentiful, whereas a state-level reading can separate the easy and the hard regions of the same task. \citet{vlacontinual} studies the \emph{sequential} regime and finds little to no forgetting from plain sequential fine-tuning with low-rank adaptation; its stability is reported as arising partly from parameter-efficient adaptation, whereas the action expert is fully trainable here, and the regime it measures is the one Section~\ref{sec:limitations} states this paper does not measure. \citet{tasknegation} audits per-skill outcomes in a multitask VLA and reports global collapse on two of ten LIBERO-Goal skills; the intervention is weight arithmetic rather than continued updates, so it is independent evidence that per-task outcomes separate where an aggregate does not, and not a competing measurement of the quantity here.

\textbf{Confidence signals read from the policy itself.} A line of work outside RL fine-tuning reads a policy's own internals to decide when it is about to fail. SAFE \citep{safe} learns from VLA features and predicts a single scalar for the likelihood of task failure, trained on successful and failed rollouts and evaluated on unseen tasks. FIPER \citep{fiper} names two indicators of impending failure. One fires when an observation looks unfamiliar, and it measures unfamiliarity by random network distillation over the embedding the policy itself computes; the other scores the entropy of the predicted action chunk. Conformal prediction on successful rollouts sets a threshold for each, and the alarm is raised only when the two are crossed together over a short window. One of those two is the family this paper gates on and the other is not, and the distinction matters more than the resemblance. Shared: random network distillation over the policy's own features, used in both places as the novelty signal. Not shared: FIPER's second indicator is the entropy of the predicted action chunk, a property of the action distribution, while the competence signal here is the PPO critic's value and the spread of a five-head value ensemble, critic quantities that have no counterpart in a monitor with no critic. What they drive differs as well: there the pair stops or warns a deployed policy, here it sets how much exploration noise a policy receives while it is being updated. A calibrated alarm and a noise gate can read the same signal and still be different mechanisms, and neither result transfers to the other by itself. Further out, agentic systems let a coding agent search the improvement strategy, analysing logs and editing the training code inside a real-world reset, execute, verify, refine loop \citep{enpire}; that is a harness around the training loop, where what we study is a mechanism inside the update.

\section{Controller: rationale, constraints and what ran}
\label{app:controller}
Success rate is not the driver, and this is a design choice rather than an omission. The two signals also differ in grain and in timing: success rate is a task-level average that is known only when an episode ends, so it lags the policy it describes. The competence signal is read from the two critics at every environment step, at the state the policy is in, so it is a state-level reading available while the episode runs. A noise scale driven by success rate goes all in on the tasks where success can be raised, because they return the most reward signal and are therefore expected to dominate the gradient, while the hard tasks return mostly penalties, and a penalty teaches only one more thing not to do out of an unbounded set. Without a regulator the policy learns the tasks that already pay and stops attending to the ones that do not, which is how competence is lost indirectly. The aim here is the healthiest exploration we can construct: a policy spread across its tasks, that knows when to explore less as well as when to explore more, and that does not chase the last few percent on a task it already performs at the cost of tasks it is losing.

The gate is a rule table rather than a learned network because a learned gate needs a training signal. Trained on return it reproduces a return-driven noise network, which is the R arm; trained on confidence it is circular, because the confidence is the quantity the mechanism is meant to construct. A fixed rule over normalized signals breaks the circle, at the price of a design that has to be argued cell by cell, which is what the reading of the four cells near the end of this appendix does.

Four things are absent from the design because they are absent from the world the policy is meant to work in. No task identity: a controller that indexed anything by task would need an enumeration of tasks, and a robot in the world meets tasks that exist nowhere else and were never named. No episode boundary: a policy that knows where an episode ends can learn to rush toward it. No timeout: length-derived signals are simulator artifacts, and a mechanism that depends on them fails at the sim-to-real boundary. No task-indexed or episode-indexed state: the controller keeps nothing that is addressed by a task, an episode or a wall-clock position. It is not memoryless, and the form of memory it does have is given in full below rather than glossed: the running statistics behind its signals are exponential moving averages held once for the whole pool, the scale computed at step $t$ is the one applied at $t+1$, and the novelty predictor is trained online so that familiarity accumulates with exposure. What is absent is history a real robot could not hold, not history as such.

The noise scale matters most where contact does. Noise while the arm moves through free space costs little; noise at the moment of a grasp costs the grasp and the time to recover it. The behavior we want from a noise gate is caution in contact-rich phases and freedom in transport phases; the controller in this paper does not see phases, and this is stated as design intent, not as a measured property.

\eqref{eq:sigma} is not decorative: the scale computed at step $t$ is used at step $t+1$, and at
$t = 0$ the base scale is returned, so the controller acts with a one-step delay. The running
statistics behind $z_i$, $v_i$ and $s_i$ are exponential moving averages held once for the whole
pool rather than per slot, and they are updated from all slots at every step. The novelty score is
not a frozen detector: the random-network target is fixed, while the predictor is trained online
against it, four Adam epochs on each iteration's own buffer at a learning rate of $10^{-3}$, after the
advantage computation and independently of the policy update. Familiarity therefore accumulates
with exposure rather than being calibrated by hand. The competence signal is gated off for outer iterations 0 through 7 and is live from 8, as Eq.~\eqref{eq:lowcomp} states, because the critic readings are not yet informative before that.

\textbf{Two critics, and which one PPO uses.} The policy carries one value head, an MLP with hidden
widths 512 and 128 on the 720-dimensional output of the action expert averaged over its tokens, with
that input detached so that no value gradient reaches the policy. It is the only critic PPO trains
and reads. Its values alone enter the truncation compensation and the generalized advantage
estimate, and it is fitted with the clipped Huber value loss of Section~\ref{sec:method} by AdamW at a
learning rate of $10^{-4}$. The controller holds a separate set of five value heads of the same
shape on the same pooled features, stored without gradient. They are trained once per outer
iteration, after the advantage computation and independently of the policy update, by four epochs of
mean-squared regression onto the returns that computation produced, with their own Adam optimizer
at $3\times10^{-4}$. Nothing they output enters an advantage, a return or the PPO loss, so no
aggregate of the five, mean or minimum, is formed in the update; the one quantity taken from them is
their standard deviation, the spread behind $s_i$. The two critics are coupled in one direction
only: the ensemble's regression targets are computed with the PPO critic's values, and no gradient
passes between them. Diversity across the five heads has two sources, a separate initialization
seed per head and a separate Bernoulli mask with probability $0.5$ that selects the samples each head
is fitted to; random priors are off. The trainer's default ensemble size is one, under which the
spread becomes half the absolute difference between the PPO critic and a single controller head.
The campaign launcher in the archive, \texttt{configs/pod\_arm\_campaign2.sh}, passes
\texttt{--ensemble\_n 5} to every arm, it takes effect only where the controller runs, and the
archived controller run configurations record \texttt{ensemble\_n} as 5.

\textbf{How the readings are standardized.} The spread at a state is the sample standard deviation
of the five heads' values there, with the $n-1$ divisor; it is neither a variance nor a range. Each
reading $x$ is standardized as $(x-\mu)/\sigma$ against an exponential moving mean $\mu$ and variance
$\sigma^2$ with decay \num{0.99}, updated one sample at a time, so the effective window is about a
hundred samples, some seven environment steps of the fourteen slots. The value and spread statistics
take in the current step's samples before standardizing them. The novelty score is standardized
against the statistics of earlier steps and enters them only afterwards, capped at five standard
deviations above their mean, so that a burst of novel states cannot widen the scale that is meant to
flag it.

\textbf{The novelty network.} Random network distillation here is a pair of MLPs of the same
shape, each two linear layers with a ReLU between them, 720 inputs, 256 hidden units and 128
outputs: a target frozen at its random initialization, and a predictor trained to match it. The
input is the 720-dimensional hidden state the action expert produces at the last denoising step,
read by a forward pre-hook at the input of the expert's action output projection, so the flow head
itself is not modified, and averaged over the fifty tokens of the action chunk; it is the same
vector the controller's value heads read. It is whitened per dimension with an exponential moving
mean and variance and clipped to $[-5, 5]$, and the raw score is the mean squared difference
between predictor and target over the 128 outputs. The predictor is trained once per outer
iteration, after the advantage computation and before the policy update: four full-batch Adam
steps at a learning rate of $10^{-3}$ on the same squared difference, over every feature vector
that iteration's buffer collected, whitened with the statistics current at that point. It has its
own optimizer and reads features stored without gradient, so its training is independent of the
policy update and changes no policy or critic weight.

\textbf{Where the thresholds come from.} On a standard normal, $1.28$ is the 90th percentile
($\Phi(1.28) \approx 0.90$), so the novelty flag marks roughly the most novel tenth of states to the
extent that the standardized score is close to normal. The shipped copy of the controller,
\texttt{code/sigma\_controller.py}, carries $1.28$ as the constructor default of its novelty threshold \texttt{z\_novel} and flags a state as novel when its standardized score exceeds that threshold; its design specification asks for a running-quantile flag.
The competence thresholds, $-0.5$ on the standardized value and $1.0$ on the standardized spread, are the defaults \texttt{v\_lo} and \texttt{s\_hi} of the same constructor, and
carry no stated derivation in the source or in the design record. All three were set when
the controller was first written and are unchanged in its source history since then. The trainer
exposes no flag for any of them and passes none, so none of the three was tuned for the reported
runs.

\textbf{On the word ``task''.} The scale is indexed by environment slot, not by task identity, and
the controller reads no task label anywhere: the word does not occur in its code, which the
supplementary archive carries as \texttt{code/sigma\_controller.py}; the two occurrences in the
original source are comments, and the shipped copy carries no comment.
In LIBERO the binding between slot and task is fixed at environment construction, so a
slot corresponds indirectly to a task for the duration of a run, but the mechanism neither knows
this nor keeps any per-task memory, and the statistics it thresholds against are pooled across
slots. This is a property of the code and is read from it rather than asserted.

\textbf{Scope of the evaluated controller.} The reported effect is obtained with two signal families and a four-cell rule; additional drivers present in the code were inactive, so the results apply to the evaluated rule. Two further drivers exist in the same source and
were switched off in these runs: an action-conditional-entropy driver, which is computed and logged
but does not move $\sigma$, and a value-trend multiplier, which defaults to $1.0$. A sixteen-cell
variant of $M$ also exists and was not used. Equations \eqref{eq:novel} to \eqref{eq:sigma} are
what the reported arms ran. The full design, with the action-conditional-entropy signal, the value
trend and the sixteen-cell table, was not ablated in this work, and no result here should be read as
evidence about it.

We build it from two readings that carry different information and that we keep apart. Novelty asks whether the policy has seen this kind of state before, and is read from a random-network-distillation score \citep{rnd}. Competence asks whether the policy can currently succeed here, and is read from two critics: the PPO critic's value for expected return, and the spread of a five-head ensemble for the disagreement between heads. A single scalar could be built from both, but a table over their combinations keeps the mechanism legible and lets the rule for each combination be stated and questioned.

In the robot's terms the four cells read as follows. Familiar and competent: it has learned this, so it should not explore, and spending updates to perfect it costs capacity that other tasks need. Novel and competent: it has not seen this situation but succeeds, so little exploration is needed. Novel and incompetent: it is meeting a new problem, it is allowed to make mistakes, and it should explore until the situation becomes familiar. Familiar and incompetent: it knows this task and still fails, so the task is hard for it, and the design calls for caution rather than exploration, consolidating what works and learning slowly. Two rationales for this last cell arose during the design, and the code runs one of them. The lower-exploration rationale is the four-state table of the project's design notes, which gives a familiar and hard state little exploration, because more exploration does not help there and risks thrashing and losing the skill; the sixteen-cell variant in the archive keeps that direction when the state is otherwise calm, setting $0.7$ for a familiar, not competent state whose other two signals are calm, and raises the same state to $1.0$ when one of those two signals fires and to $1.3$ when both do. The higher-exploration rationale is the literature synthesis behind the controller's design specification, which gives a familiar state with low competence a moderate scale, because the policy is in distribution but struggling and modest exploration is warranted. The code follows the second: $1.4$ where that synthesis asks for the highest scale, $1.0$ where it asks for the default, $0.4$ where it asks for the lowest, and $1.2$ in this cell, slightly above the base. Against the four-state table the running rule therefore does the opposite in this cell, raising the scale where that design lowers it, and every result in this paper is for the running rule. In the design, the novelty signal exists to separate exactly the case in which the competence signal alone would mislead: low competence on a novel state calls for exploration, and low competence on a familiar state does not. That is why the design's table is split over two signals and cannot be reduced to a single rule that explores wherever competence is low.

The competence signal is held off for the first eight iterations because the critic readings are not informative before their warm-up; the count is read from the campaign launch banner, which prints \texttt{critic\_warmup=8}, and it was not tuned. The base scale $\sigma_0 = 0.3$ and the clip $[0.1, 0.7]$ are read from the campaign controller launch banner, \texttt{sigma base=0.3 clamp[0.1,0.7]}, and were fixed before the campaign rather than tuned during it. The intent behind them is stated as design and not as measurement: the reachable band is meant to stay above the level at which the injected noise stops changing behaviour and below the level at which the sampler's actions lose coherence. The range $[0.12, 0.42]$ of the four values listed in Section~\ref{sec:method} is not a third setting but the span the table can reach from that base, and it lies inside the clip, so the clip never binds on a value the table produces. The design rationale in full, the real-world constraints behind it, the one-step lag, the running statistics and the code-level proof that no task label is read are set out earlier in this appendix.

\section{Collapse definitions}
\label{app:definitions}
All four definitions share one threshold, one half of the same task's tare on the same instrument. \emph{Pooled} averages the last five canonical scans (150 episodes per task) and then thresholds; it is the primary definition. \emph{Point-wise} thresholds a single 30-episode scan and is the definition used for the survival analysis. \emph{Sustained} counts a task that stayed below the threshold on two consecutive scans and is a secondary definition. \emph{Ever} counts a task that crossed at any scan and never shrinks. The four disagree exactly where a task crossed and later recovered, which the pooled column shows happening. Task identities and first-crossing iterations for every run are in \texttt{t5\_collapse\_definitions.md}.

The quantity resembles the forgetting measures of continual learning, which characterize a model by transfer across tasks rather than by final accuracy alone \citep{gem} and pair forgetting with intransigence \citep{rwalk}, and we borrow the habit of measuring per task rather than in aggregate. It is not backward transfer. Those measures are defined over a \emph{sequential} stream, where a task is left behind and revisited; our seven tasks are trained simultaneously throughout, so what we measure is interference within one joint objective, and a task that falls can also recover, which our pooled column shows happening. We report it under continued training on the same task set, a setting a robotic continual-RL benchmark argues has been crowded out by attention to catastrophic forgetting \citep{continualworld}. All four definitions are stated at the head of this appendix and side by side in Table~\ref{tab:definitions}.

At the seed level a Fisher exact test on the pooled definition gives $p = 1.000$ for fixed against learned (2/3 against 3/3) and $p = 0.100$ for controller against learned (0/3 against 3/3); the test of fixed noise against the controller, and the resolution limit of three seeds per arm, are stated in Section~\ref{sec:collapse}. All three tests are computed by \texttt{stat\_audit.py} from the same routine that prints the table. Our design uses three independent training seeds per arm, so we report the exact results alongside the observed collapse counts and trajectories. The evidence is at the level of the task-seed pair, of which there are \num{63}, seven tasks on each of the nine complete seeds of the three arms, \num{31} of them with a first-crossing iteration and \num{32} censored at iteration 200 with no crossing, and in the survival curves that use them. Those \num{63} pairs are not \num{63} independent training runs and are not treated as replicates: the seven tasks of one seed share a single training trajectory, a single set of weights and a single optimizer history, so they are correlated by construction. The independent unit of this design is the training seed, and there are three per arm. Two cases show why we report all four rather than choosing. Controller seed 3337 crosses on t4 at outer 160, recovers it at 180 and is below again on its last scan at 200, so it counts under the last-scan and ever definitions and not under the sustained one, which asks for two consecutive scans. Read from \texttt{evaluation/canon/k\_s3337}, that task's rate against its threshold of \num{0.2} is \num{0.167} at 160, \num{0.333} at 180 and \num{0.167} at 200, and is at or above \num{0.233} at every earlier scan. Fixed seed 3337 is the mirror image: it holds t0 below the threshold across the scans at outer 80, 100 and 120, then recovers it, so it counts as sustained while both its pooled and its last-scan counts are zero. Read from \texttt{evaluation/canon/f\_s3337}, that task's rate is \num{0.267}, \num{0.233} and \num{0.233} against its threshold of \num{0.3}, and it is above the threshold at every other scan. Every campaign run is scanned every 20 iterations and pools the same window, so the sustained column is read across arms at one cadence. The one exception is outside the campaign: the local-instrument controller run has a \num{14}-episode fragment at iteration 160, which the \num{210}-episode completeness rule excludes, so its five complete scans span \num{100} to \num{200}, and that row is already marked not comparable to the campaign rows. All four columns, with the task identities and the iteration of first sustained crossing, are in \texttt{t5\_collapse\_definitions.md}. Table~\ref{tab:definitions} typesets the four columns for every run; the task identities behind each count stay in that file, which is released with the measurement tools.

\begin{table}[!ht]
\centering
\small
\setlength{\tabcolsep}{3pt}
\begin{tabular}{lrrrrrr}
\toprule
arm & last & pooled & last scan & sustained & ever & first crossing \\
\midrule
K controller, seed 2337 & \num{200} & \num{0/7} & \num{0/7} & \num{0/7} & \num{0/7} & none \\
K controller, seed 3337 & \num{200} & \num{0/7} & \num{1/7} & \num{0/7} & \num{1/7} & \num{160} \\
F fixed 0.25, seed 1337 & \num{200} & \num{4/7} & \num{7/7} & \num{6/7} & \num{7/7} & \num{20} \\
F fixed 0.25, seed 2337 & \num{200} & \num{1/7} & \num{4/7} & \num{2/7} & \num{4/7} & \num{120} \\
F fixed 0.18, seed 1337 & \num{200} & \num{1/7} & \num{2/7} & \num{1/7} & \num{3/7} & \num{60} \\
F fixed 0.25, seed 3337 & \num{200} & \num{0/7} & \num{0/7} & \num{1/7} & \num{2/7} & \num{80} \\
F fixed 0.15, seed 1337 & \num{200} & \num{0/7} & \num{0/7} & \num{1/7} & \num{1/7} & \num{120} \\
R learned, seed 1337 & \num{200} & \num{4/7} & \num{5/7} & \num{4/7} & \num{5/7} & \num{40} \\
R learned, seed 2337 & \num{200} & \num{6/7} & \num{6/7} & \num{6/7} & \num{6/7} & \num{20} \\
K controller, seed 1337, local instrument & \num{200} & \num{1/7} & \num{3/7} & \num{2/7} & \num{3/7} & \num{80} \\
K controller, seed 1337 & \num{200} & \num{0/7} & \num{0/7} & \num{0/7} & \num{0/7} & none \\
K controller, seed 1337, fp32 master copy & \num{200} & \num{3/7} & \num{5/7} & \num{3/7} & \num{7/7} & \num{60} \\
F fixed 0.25, seed 1337, fp32 master copy & \num{200} & \num{7/7} & \num{7/7} & \num{7/7} & \num{7/7} & \num{20} \\
F fixed 0.15, seed 2337 & \num{200} & \num{2/7} & \num{3/7} & \num{2/7} & \num{4/7} & \num{20} \\
R learned, seed 3337 & \num{200} & \num{5/7} & \num{5/7} & \num{5/7} & \num{6/7} & \num{40} \\
\bottomrule
\end{tabular}
\caption{The four collapse definitions side by side, generated from the same producer that writes the owner file. All four share one threshold, half of each task's own tare on the same instrument, and differ only in what they apply it to: \emph{pooled} over the last five complete scans, \emph{last scan} over one, \emph{sustained} over two consecutive, \emph{ever} cumulative. \emph{Last} is the final complete scan and \emph{first crossing} the earliest iteration at which any task crossed the threshold on any scan, which is the first event of the \emph{ever} column and not of the \emph{sustained} one. A row can therefore read \num{0/7} sustained and still carry a first crossing, as the seed-3337 controller row does at \num{160}: that task crossed once and did not cross on the following scan. The owner file carries a separate first-sustained column, which for that row reads none; it is not printed here. \emph{None} appears where no crossing occurs at all. Counts are tasks out of seven. Every row reached iteration 200 and was scanned every 20 iterations, so the three columns that count opportunities rather than time are read across rows at one cadence; runs that did not reach it enter no count in this paper and stay in the released measurement files.}
\label{tab:definitions}
\end{table}

\section{Per-task and per-seed results}
\label{app:pertask}
\begin{figure}[h]
\centering
\includegraphics[width=\textwidth]{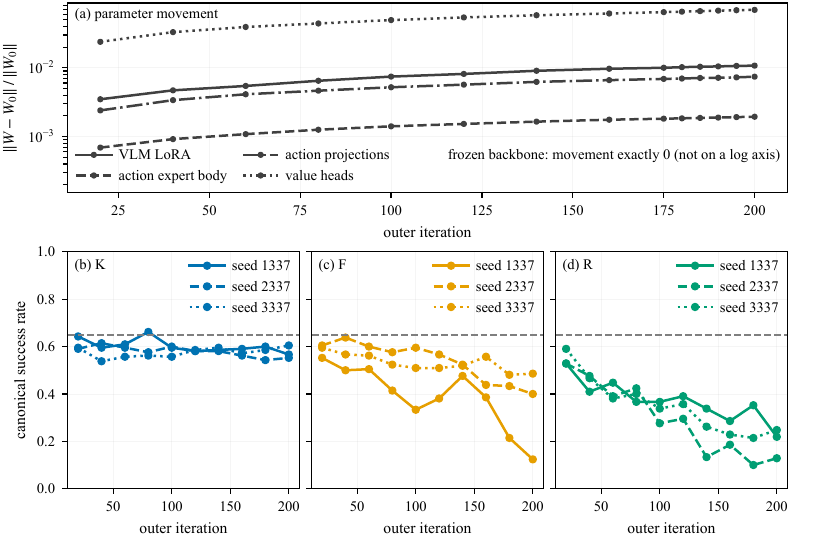}
\caption{(a) Relative Frobenius movement of each trainable parameter group from the outer-0 baseline, $\lVert W_t - W_0 \rVert_F / \lVert W_0 \rVert_F$ with $W_t$ the group's weights at outer iteration $t$, which the axis label writes as $W$; Table~\ref{tab:movement} reports the same quantity over 40-iteration windows. The panel is read from 15 sha256-verified checkpoint packages of controller seed 2337 and drawn on a logarithmic axis. Each group is normalised by its own norm, so magnitudes across groups are not a ranking, and the frozen backbone is named in the legend but not drawn because its relative movement is exactly 0, which has no position on a logarithmic axis. (b) to (d) the same canonical curves as Fig.~\ref{fig:curves}, grouped so that the seeds of one arm are read against each other, with each arm's behavior-cloning tare as the grey dashed line. Two rulers apply: the Wilson 95\% interval of a single 210-episode scan (half-width about 0.067) and the run-to-run rescoring SD of 0.0375 on the campaign instrument.}
\label{fig:param}
\label{fig:seeds}
\end{figure}

\subsection{Noise allocation across tasks: how it was read}
\label{app:sigmaalloc}
\textbf{What the log contains, and what it therefore cannot answer.} Each training log writes one line per outer iteration. That line carries a mean applied scale and a seven-element field holding one scale per task, already aggregated over the 14 parallel environments. Step-level allocation is not recorded anywhere, so how a scale was distributed \emph{within} an outer iteration cannot be recovered from these files and is not claimed. The novelty and competence readings on the same line are per-iteration scalars, not per task, so the signal that drove a per-task value is likewise not separable per task from this record.

\textbf{The two fields are not assumed to be one quantity.} The logged mean and the unweighted mean of the seven per-task values were compared outer by outer before either was used in a sentence. On the fixed arm they are identical to four decimals, as a fixed scale requires. On the controller they differ by at most \num{0.0033} on any single iteration and by at most \num{0.0002} in the run mean. The mean applied noise quoted in Section~\ref{sec:analysis} is the logged mean field; the per-task figures are the per-task field; the two are never averaged together.

\textbf{How the groups were formed.} The seven tasks were ordered by their own behavior-cloning tare on the campaign instrument, measured from the 210-episode canonical scan, and split into the three lowest and the three highest. The lower group is t5 (\num{0.367}), t4 (\num{0.400}) and t0 (\num{0.600}); the higher group is t3 (\num{0.800}), t2 (\num{0.833}) and t1 (\num{0.900}). \textbf{The seventh task is t6}, KITCHEN\_SCENE6, tare \num{0.633}, which sits in the middle of the order and belongs to neither group. It is excluded from the comparison and not from the paper: its own mean scale is \num{0.307} on seed 1337 and \num{0.270} on seed 2337, that is, near the top of the range rather than in the middle of it, which is the clearest evidence in this table that tare order and allocation order are not the same order. The rank correlation between tare and mean allocated scale is $-0.857$ on seed 1337 and $-0.929$ on seed 2337 over seven points. These correlations describe the seven evaluated tasks; we do not use them to make an inferential claim about a wider task population.

\textbf{Weighting.} Every mean here is unweighted: the mean over the covered outer iterations of a task, then the mean over the three tasks of a group. Iterations are not weighted by episode count, and tasks are not weighted by tare.

\textbf{Missing iterations, and what is not done about them.} Seed 1337 has a per-task line for all 200 outer iterations. Seed 2337 has one for 155 of them, o0 to o154, with 45 missing at the end of the run; the figures for that seed are means over the range it covers and are not a reading of a 200-iteration run. Nothing is interpolated across a missing iteration. Where a resumed run wrote more than one line for the same iteration index, the last line for that index is kept. The corresponding log for seed 3337 is not present in the repository or in the checkpoint archive, so that seed contributes nothing to this table.

\textbf{The task label.} The controller reads task-agnostic signals and receives no task label at any point. The labels above are used only to group the logged output after the fact, so that allocation can be read against difficulty. Grouping a record by a variable the mechanism never saw is a description of the record, not a description of the mechanism.

\textbf{What this measures.} It measures allocation. The arms were varied as wholes, so this table does not isolate how much of the controller's effect is due to distributing noise across task streams as opposed to adapting it over training time or to the coupling between the two. Producer: \texttt{sigma\_vs\_tare.py} and \texttt{sigma\_campaign\_coverage.py}.

\textbf{Parameter movement.} Over 200 iterations of controller seed 2337 the action expert moves by \num{0.19}\% of its weight norm relative to the BC weights, the LoRA adapters by \num{1.1}\%, the value heads by \num{7}\%; the frozen backbone moves by exactly zero in every checkpoint (null control). Movement is diffusive rather than directed (Fig.~\ref{fig:param}a). That seed loses no task under any of the four definitions (Table~\ref{tab:pool}), so it does not by itself place collapse at a small displacement. The runs that do collapse are measured over 40-iteration windows (Table~\ref{tab:movement}): in bfloat16 the action expert of the fixed and learned arms moves by \num{0.0008} to \num{0.0010} of its norm per window, and that of the controller by \num{0.0008} to \num{0.0009}. In this regime tasks are lost and kept at action-expert displacements of the same order, a small fraction of a percent of its norm per window.

Fixed noise is seed-dependent: at $\sigma=0.25$ the pooled plateau spans \num{0.3162}, \num{0.4724} and \num{0.5105} across seeds 1337, 2337 and 3337, all three measured to iteration 200, and the first threshold crossing occurs at iteration 20, 120 and 80 respectively. The spread across those three seeds, \num{0.194}, is more than five times the run-to-run rescoring SD, so it is a property of the seeds and not of the instrument. The three controller seeds pool at \num{0.585}, \num{0.589} and \num{0.564}, a standard deviation of \num{0.013} inside the run-to-run rescoring SD of \num{0.0375}, with the widest pair \num{0.0248} apart; one of the three was continued from its iteration-180 checkpoint to reach 200, and what the continuation did to its reading is not measured. The per-seed curves are in Fig.~\ref{fig:seeds}.

The only timing the run record contains is the rollout duration of each iteration, and it is wall-clock time from separate rented nodes and a laptop, so it cannot be read as an algorithmic cost difference between the arms and no conclusion about their relative cost is drawn from it; Appendix~\ref{app:reinflow} uses wall-clock time only to set the order of magnitude of one benchmark's iteration against another's, where the gap is several fold rather than a few percent.

\section{Precision: configuration evidence and the fp32 pilot}
\label{app:precision}
Mixed-precision training keeps an fp32 master copy so that small updates are not rounded away \citep{mixedprecision}. The reference VLA RL recipes we ported train the policy in bfloat16 in place \citep{rlinf,openpi}: in the pinned RLinf clone the LIBERO PPO configuration sets \texttt{precision: null} for the model and for the FSDP parameter, reduce and buffer dtypes, which disables FSDP mixed precision, with autocast and the gradient scaler both off, and the optimizer constructed directly as AdamW (config file and line numbers recorded in the reproducibility appendix). The reference paper itself carries no precision row at all: its hyperparameter tables list no dtype, no master-copy, no optimizer-state precision and no gradient-accumulation entry, so the recipe is recoverable only from code, and in that code the FSDP block ties \texttt{param\_dtype}, \texttt{reduce\_dtype} and \texttt{buffer\_dtype} to one variable rather than using the usual bfloat16-compute with fp32-reduce split. In language-model RL the same rounding question has been measured directly: bfloat16 carries seven mantissa bits against fp16's ten, moving to fp16 shrinks the training-inference mismatch by about a factor of 24, and bf16 GRPO runs collapse early at 73 to 84 percent peak accuracy \citep{fp16mismatch}. We found no equivalent measurement for a robot or vision-language-action policy. We measure the consequence in Section~\ref{sec:analysis}.

The direction agrees with what is measured in language-model RL, where bfloat16 rounding is identified as the root cause of a training-inference mismatch and fp16 removes it \citep{fp16mismatch}; the magnitude below is ours, because we found no such measurement for a flow VLA. A pilot with an fp32 master copy has been measured to iteration 200 and its readings are in \texttt{t6\_fp32\_pilot.md}. The pilot shows three patterns. First, the rounding was acting as a brake rather than only as a defect, and the gap widens with training: over the ten canonical scans the controller seed has at both precisions, it pools at \num{0.603} \num{[0.582, 0.624]} in bfloat16 against \num{0.388} \num{[0.367, 0.409]} with an fp32 master, non-overlapping, and the same comparison reads \num{0.616} against \num{0.516} over the first three scans and \num{0.586} against \num{0.243} over the last three. The fixed arm, matched over all ten of its scans, reads \num{0.389} against \num{0.219}. Second, both pilot arms crossed the collapse threshold at the reference learning rate, the controller on two tasks by iteration 60 and the fixed arm on one by iteration 20, so removing the rounding does not remove the collapse; it lets the learning rate bite. Third, at equal exposure, both arms being measured to iteration 200, the fp32 controller arm carries fewer collapsed tasks than the fixed arm under the pooled and sustained definitions, \num{3/7} against \num{7/7} in each, and on the last scan, \num{5/7} against \num{7/7}; under the ever definition the two arms are level at \num{7/7}, so the ordering holds on three of the four definitions and not on all four. Each pilot arm uses one seed. The controller pair's run configurations were compared field by field and differ in the fp32 master flag and two directory names, while the two processes ran from different code commits; the fixed pair has no run configuration on record (Appendix~\ref{app:repro}). The pre-registered questions are in \texttt{prereg\_fp32\_pilot.md}, and the pilot holds the learning rate fixed by design.

\begin{figure}[h]
\centering
\includegraphics[width=\textwidth]{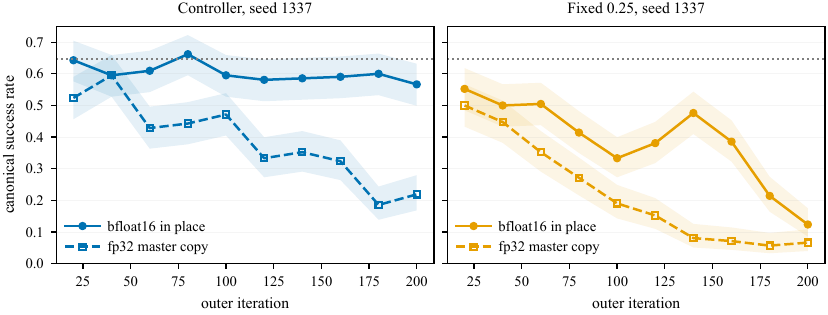}
\caption{Canonical success rate against outer iteration for two arms in two numerical regimes. Solid: the campaign regime, bfloat16 weights updated in place. Dashed: the same arm with an fp32 master copy. Shaded bands are the Wilson 95\% interval of each 210-episode scan; the grey dotted line is the behavior-cloning tare on the campaign instrument, \num{0.6476}. Each series is a single seed, 1337, at the reference actor learning rate. Produced by \texttt{make\_figures.py}.}
\label{fig:fp32}
\end{figure}

\begin{table}[h]
\centering
\small
\setlength{\tabcolsep}{4pt}
\begin{tabular}{llrlll}
\toprule
arm & precision & last scan & pooled & Wilson 95\% & first crossing \\
\midrule
K, seed 1337 & bfloat16 in place & 200 & \num{614/1050} = \num{0.5848} & \num{[0.5547, 0.6142]} & none \\
K, seed 1337 & fp32 master copy & 200 & \num{297/1050} = \num{0.2829} & \num{[0.2564, 0.3109]} & iteration 60 \\
F 0.25, seed 1337 & bfloat16 in place & 200 & \num{332/1050} = \num{0.3162} & \num{[0.2888, 0.3449]} & iteration 20 \\
F 0.25, seed 1337 & fp32 master copy & 200 & \num{90/1050} = \num{0.0857} & \num{[0.0703, 0.1042]} & iteration 20 \\
\bottomrule
\end{tabular}
\caption{Pooled canonical success over the last five canonical scans of each series, 1050 episodes per row, against the campaign tare of \num{0.6476}. The run-to-run rescoring SD on this instrument is \num{0.0375} and the pooled Wilson half-width is about \num{0.030}, so the separation between the two precisions of the same arm is about eight times the repeat spread in the controller family and six times it in the fixed family. The first-crossing column gives the earliest iteration at which any task fell below half of its own tare value. Each row is a single seed. Supplementary files \texttt{t1\_plateau\_pool.md} and \texttt{t6\_fp32\_pilot.md}.}
\label{tab:pilot}
\end{table}

\begin{table}[h]
\centering
\small
\begin{tabular}{lrrrr}
\toprule
group & elements & unchanged & unchanged \% & tensors that moved \\
\midrule
frozen side (VLM body) & \num{350165184} & \num{350165184} & \num{100.00} & \num{0/345} \\
action expert body & \num{98245840} & \num{94338243} & \num{96.02} & \num{112/145} \\
flow head & \num{1635152} & \num{54} & \num{0.00} & \num{10/10} \\
VLM LoRA & \num{819200} & \num{496667} & \num{60.63} & \num{124/128} \\
value head & \num{434944} & \num{1} & \num{0.00} & \num{5/5} \\
\midrule
total & \num{451300320} & \num{445000149} & \num{98.60} & \\
\bottomrule
\end{tabular}
\caption{Element-level change across three outer iterations of \texttt{k\_corrected\_s1337}, \texttt{iter\_000033} against \texttt{iter\_000036}. This is an earlier seed-1337 controller run, not one of the released campaign arms; the Reproducibility Statement describes it, and its run configuration is in the archive. \num{96.02}\% of the action expert's elements are bit-identical while \num{112} of its \num{145} tensors move in at least one element. A tensor counts as moved when any one of its elements changes, so the two counts do not contradict each other; whether an unchanged element received an update below bfloat16 resolution, or none, was not measured. The frozen side is bit-identical in \num{100}\% of its elements, which is the null control that shows the comparison works. This counts \emph{elements} over three iterations and Table~\ref{tab:movement} counts \emph{tensors} over forty; neither bounds the other. Every count is over distinct tensors. The checkpoint's \texttt{model\_state\_dict} holds \num{1261} entries because the flow model is stored under two names, \texttt{policy.model.*} and \texttt{flow\_model.*}: all \num{628} pairs are equal in both checkpoints and no pair moves differently, so an entry count is twice the distinct count except for the value head, which is stored once. Counted once, the policy holds \num{451300320} parameters, the published size of \num{450}M. The group percentages are the same under either count; the total's is \num{98.60} counted once and \num{98.65} over entries. The counts are from supplementary file \texttt{bf16\_swallowed\_updates.md}, an authored record of readings taken by comparing the two checkpoints' \texttt{model\_state\_dict} element by element. The arithmetic beside them is measured by the shipped script \texttt{test\_fp32\_master\_swallow.py}: a weight of magnitude \num{0.02} sits $1.221 \times 10^{-4}$ from its nearest bfloat16 neighbour, and twenty consecutive updates of $5 \times 10^{-6}$ move it by \num{0} in place while accumulating to $9.999 \times 10^{-5}$ on an fp32 master copy.}
\label{tab:round}
\end{table}

\subsection{Parameter movement, read from the weights}
\label{app:movement}

The precision question above asks how much of the trainable surface the optimizer actually moves. This table answers it from the weights rather than from the configuration. Each row opens the two checkpoints at the ends of one window, both verified by sha256, and reports two quantities per group of trainable tensors: relative movement, $\lVert W_{t_1} - W_{t_0} \rVert_F / \lVert W_{t_0} \rVert_F$, where $W_{t_0}$ and $W_{t_1}$ are the group's weights at the first and the last outer iteration of the window, and the fraction of tensors that are bit-identical under \texttt{torch.equal}. The producer is \texttt{measure\_param\_movement.py} and the supplementary file is \texttt{parameter\_movement\_by\_arm.md}, from which every cell below is parsed rather than retyped.

The table shows two descriptive findings within the measured windows. The bit-identical fraction separates the two numerical regimes cleanly: \num{11.9} to \num{13.3}\% of trainable tensors never move at all under bfloat16, against \num{1.4}\% in the fp32-master pilot. And within bfloat16 the ordering across groups is stable: the value heads move most, the LoRA adapters next, and the action expert least by roughly a factor of five, while under an fp32 master the expert closes most of that gap. A third reading follows from all three exploration arms being in the table: the action expert's displacement lies between \num{0.0008} and \num{0.0010} across every bfloat16 window of every arm, and the fixed arm reads the same expert movement at $\sigma = 0.25$ as at $0.15$. Within these windows, then, the action expert of every arm keeps changing by a comparable amount, and task preservation in the controller arm coincides with that continued change rather than with its absence. A comparable displacement norm is not evidence that the update dynamics are otherwise alike, and no such conclusion is drawn from it. The bfloat16 and fp32 rows come from different runs, compared field by field only for the controller pair and differing there in more than the precision (Appendix~\ref{app:repro}), so the contrast is an observation about two regimes and not a controlled experiment.

Three limits are load-bearing. First, this is a different quantity from the \num{96.02}\% quoted in Section~\ref{sec:analysis}: that is a fraction of \emph{elements} across three consecutive iterations, this is a fraction of \emph{tensors} across a 40-iteration window, and neither bounds the other. Second, relative movement is a magnitude and not an improvement; a larger number here says only that more of the surface changed. Third, the supplementary file opens \num{39} of \num{63} cells and the rest are unmeasured rather than zero. The reason is not distance but absence or duplication, measured over every arm that owns a cell and sorted into four classes that the supplementary file names cell by cell. In \num{13} cells the window starts at iteration 0 and the run never wrote one: the fixed and learned arms save after every fifth outer iteration and their first file is iteration 5, so iteration 0 exists only as a rebuilt file, and this table is read from files the runs wrote. In \num{5} cells the run stopped before the far end of the window, at outer \num{2} for a resume-and-reset smoke directory and outer \num{75} for a fixed-noise ablation. In \num{4} cells the directory that owns the cell is a second name for a run that is measured under its first: the two names carry the same checksum at every iteration both of them hold, the short-named copy does not hold this window's end, and the window itself is measured on the twin's row. In \num{2} cells the endpoint was written and then deleted by the trainer's keep-two pruning before it was copied, which is the seed-3337 learned arm's declared gap. Spending more compute recovers none of them: three of the classes are files that exist nowhere, and the fourth is already read. A separate defect the sweep did find was on the measuring side, where a checksum flag was read as a verification and a zero-length file was accepted as a window end; the flag is not a verification and the check refuses zero length, a wrong digest and a missing marker.

The iteration-0 end of the fixed and learned arms is a third case, and it is neither a loss nor a measurement. Those runs write a checkpoint after every fifth outer iteration, so iteration 0 was never a file; it can be rebuilt, because the trainer constructs the model from the behavior-cloning weights and the run's own seed before any arm-specific code runs. Rebuilt and compared against the three controller runs that did write a \texttt{complete\_000000} bundle, \num{1261} of \num{1261} state-dict entries, \num{633} distinct tensors, are bit-identical for seeds 1337, 2337 and 3337, with no difference in dtype and none in value; the tool that rebuilds them is in the supplementary archive and the three files are in the weight release, so the iteration-0 end of a window is not an absent file for a reader. The fp32-master pilot is rebuilt the same way with the trainer's own promotion of the trainable tensors to fp32, which also happens before the arm branch, and against the fp32 controller pilot's own \texttt{complete\_000000} it is again \num{1261} of \num{1261} bit-identical; that file is in the release as well. The rows stay unmeasured in this table, because it is read from checkpoints the runs wrote and a rebuilt end is not a reading of one. The learned arm's nine \texttt{reinflow\_noise.*} tensors sit outside this: they do not exist at iteration 0 at all, the noise network being created at the first rollout, which is a property of the arm rather than a missing file.

\begin{table}[h]
\centering
\small
\setlength{\tabcolsep}{4pt}
\begin{tabular}{llllrrrr}
\toprule
arm & seed & variant & window & expert & LoRA & critic & bit-ident.\,\% \\
\midrule
F & 1337 & fixed 0.25 & 100--140 & \num{0.0009} & \num{0.0048} & \num{0.0175} & \num{13.3} \\
F & 1337 & fixed 0.25 & 160--200 & \num{0.0009} & \num{0.0046} & \num{0.0186} & \num{13.3} \\
F & 1337 & fp32 master & 100--140 & \num{0.0034} & \num{0.0057} & \num{0.0177} & \num{1.4} \\
F & 1337 & fp32 master & 160--200 & \num{0.0034} & \num{0.0056} & \num{0.0209} & \num{1.4} \\
F & 1337 & fixed 0.15 & 100--140 & \num{0.0009} & \num{0.0044} & \num{0.0181} & \num{13.3} \\
F & 1337 & fixed 0.15 & 160--200 & \num{0.0009} & \num{0.0046} & \num{0.0156} & \num{13.3} \\
F & 1337 & fixed 0.18 & 100--140 & \num{0.0009} & \num{0.0046} & \num{0.0179} & \num{13.3} \\
F & 1337 & fixed 0.18 & 160--200 & \num{0.0009} & \num{0.0045} & \num{0.0166} & \num{13.3} \\
F & 2337 & fixed 0.25 & 100--140 & \num{0.0009} & \num{0.0051} & \num{0.0190} & \num{13.3} \\
F & 2337 & fixed 0.25 & 160--200 & \num{0.0009} & \num{0.0047} & \num{0.0165} & \num{13.3} \\
F & 2337 & fixed 0.15 & 100--140 & \num{0.0008} & \num{0.0044} & \num{0.0171} & \num{13.3} \\
F & 2337 & fixed 0.15 & 160--200 & \num{0.0008} & \num{0.0043} & \num{0.0156} & \num{13.3} \\
F & 3337 & fixed 0.25 & 100--140 & \num{0.0009} & \num{0.0046} & \num{0.0188} & \num{13.3} \\
F & 3337 & fixed 0.25 & 160--200 & \num{0.0009} & \num{0.0045} & \num{0.0181} & \num{13.3} \\
\midrule
K & 1337 & --- & 0--40 & \num{0.0009} & \num{0.0045} & \num{0.0331} & \num{13.3} \\
K & 1337 & --- & 100--140 & \num{0.0009} & \num{0.0046} & \num{0.0163} & \num{13.3} \\
K & 1337 & --- & 160--200 & \num{0.0009} & \num{0.0044} & \num{0.0172} & \num{13.3} \\
K & 1337 & controller dir. & 0--40 & \num{0.0009} & \num{0.0044} & \num{0.0337} & \num{13.3} \\
K & 1337 & controller dir. & 100--140 & \num{0.0008} & \num{0.0044} & \num{0.0174} & \num{13.3} \\
K & 1337 & controller dir. & 160--200 & \num{0.0008} & \num{0.0045} & \num{0.0169} & \num{13.3} \\
K & 1337 & fp32 master & 0--40 & \num{0.0035} & \num{0.0056} & \num{0.0335} & \num{1.4} \\
K & 1337 & fp32 master & 100--140 & \num{0.0033} & \num{0.0055} & \num{0.0176} & \num{1.4} \\
K & 1337 & fp32 master & 160--200 & \num{0.0032} & \num{0.0050} & \num{0.0173} & \num{1.4} \\
K & 2337 & --- & 0--40 & \num{0.0009} & \num{0.0047} & \num{0.0331} & \num{13.3} \\
K & 2337 & --- & 100--140 & \num{0.0008} & \num{0.0045} & \num{0.0182} & \num{13.3} \\
K & 2337 & --- & 160--200 & \num{0.0008} & \num{0.0043} & \num{0.0168} & \num{13.3} \\
K & 3337 & --- & 0--40 & \num{0.0009} & \num{0.0047} & \num{0.0375} & \num{13.3} \\
K & 3337 & --- & 100--140 & \num{0.0008} & \num{0.0042} & \num{0.0181} & \num{13.3} \\
K & 3337 & --- & 160--200 & \num{0.0009} & \num{0.0044} & \num{0.0167} & \num{13.3} \\
\midrule
R & 1337 & --- & 100--140 & \num{0.0010} & \num{0.0050} & \num{0.0167} & \num{11.9} \\
R & 1337 & --- & 160--200 & \num{0.0010} & \num{0.0051} & \num{0.0168} & \num{11.9} \\
R & 2337 & --- & 100--140 & \num{0.0010} & \num{0.0052} & \num{0.0171} & \num{11.9} \\
R & 2337 & --- & 160--200 & \num{0.0010} & \num{0.0052} & \num{0.0165} & \num{12.6} \\
\bottomrule
\end{tabular}
\caption{Relative movement of the three trainable groups over 40-iteration windows, and the fraction of their tensors that are bit-identical across the window. Movement is $\lVert W_{t_1} - W_{t_0} \rVert_F / \lVert W_{t_0} \rVert_F$, with $W_{t_0}$ and $W_{t_1}$ a group's weights at the first and the last outer iteration of the window; it is the quantity of Figure~\ref{fig:param}a, which fixes $t_0 = 0$. The last column counts tensors, not elements. The fixed arm is measured at its campaign value $\sigma = 0.25$ on all three seeds, at the $0.15$ ablation on both of its seeds and at the $0.18$ ablation on its one; the rows marked \emph{fp32 master} are the precision pilot of Figure~\ref{fig:fp32}, one seed per arm. The controller rows marked \emph{controller dir.} are the local seed-1337 run, the one scored on the local instrument in Table~\ref{tab:pool}; its values differ from the campaign seed-1337 run in every window, so the two are separate runs and are not pooled. The supplementary file records \num{39} measured cells of \num{63}; six of them are rows of a run held under more than one directory name and are equal to the last digit in every group, so they are merged and \num{33} distinct windows are printed. Absent windows are unmeasured, not zero. Rows are emitted by \texttt{movement\_table\_tex.py}, which parses \texttt{parameter\_movement\_by\_arm.md} rather than retyping it.}
\label{tab:movement}
\end{table}

\ifreinflowdata
\clearpage
\section{Beyond the VLA: ReinFlow's benchmarks}
\label{app:reinflow}
ReinFlow \citep{reinflow} is the closest published method to ours in mechanism, since it also learns the injected noise of a flow policy, and it is the furthest from ours in what it is measured on. Its own conclusion names the gap: it reports that its experiments use relatively small networks and that scaling to large flow-based vision-language-action models remains future work. This appendix sets the two evaluation profiles beside each other so that a difference in breadth is not read as a difference in rigour, and so that the cost of our own narrowness is stated in the same units as theirs.

Three things are fixed before the table is read. First, breadth and depth trade against each other at a fixed budget. Their span is ten tasks over three benchmark families with two headline baselines, and a further appendix comparison against diffusion-RL baselines on three locomotion tasks whose names appear only inside its figures. Ours was one benchmark family and seven tasks, with three seeds per arm and two rulers in every cell. Ours is that plus \FactGymFamiliesWord{} of their ten tasks, all but Transport, over \FactPolicyClassesWord{} policy classes at the same \FactSeedsWord{} seeds, \FactGymCompleteWord{} of which completed. Second, the two are not at the same price. Their reported per-iteration figures, in \texttt{agent/eval/visualize/wall-time.md} of the pinned repository, are \num{11.6} to \num{30.9} for state input and \num{218} to \num{313} for pixel input; that file states no unit and no hardware, so reading them as seconds per iteration is an assumption. On ours an outer iteration costs \num{451} to \num{1108} seconds and a single 210-episode canonical scan \num{2572} to \num{4630}. The two are read only for their order of magnitude, not as a cost difference at a finer grain (Appendix~\ref{app:pertask}). Third, seeds: they train three seeds for every task and five on two Franka Kitchen variants, and shade their curves with one standard deviation across runs.

\begin{table}[!ht]
\centering
\small
\setlength{\tabcolsep}{4pt}
\begin{tabular}{llllll}
\toprule
benchmark family & tasks & input & baselines & seeds & error bar \\
\midrule
OpenAI Gym & \num{4} & state & DPPO, FQL & \num{3} & mean $\pm$ 1 SD \\
Franka Kitchen & \num{3} & state & DPPO, FQL & \num{3}, \num{5} on two & mean $\pm$ 1 SD \\
Robomimic & \num{3} & pixel & DPPO, FQL & \num{3} & mean $\pm$ 1 SD \\
\midrule
ours, LIBERO-10 & \num{7} & pixel and state & BC tare, three arms & \num{3} per arm & Wilson and run-to-run SD \\
ours, their gym & \num{8} of \num{9} & state, pixel & three arms & \num{3} per arm & seed SD, eval SD \\
\bottomrule
\end{tabular}
\caption{Evaluation profile of \citet{reinflow} beside ours. Cells are read from the paper's own text and tables; a string that names an environment is not counted as an experiment on it. Our first row is the campaign described in Section~\ref{sec:setup}; our second is the re-run of Appendix~\ref{app:reinflowgym}, where \FactGymFamilies{} tasks were attempted and \FactGymCompleteWord{} completed at three seeds in all three arms.}
\label{tab:reinflow}
\end{table}
\subsection{The reference mechanism re-run on its own benchmarks}
\label{app:reinflowgym}
The comparison above is between evaluation profiles. This subsection is the other half of it: the three exploration arms of this paper, the learned one following \citet{reinflow}, run on that paper's own benchmarks at matched budget, so that the mechanism is read somewhere other than the suite it was designed against here. In this subsection a family is one task of the reference together with the runs made on it, up to three seeds in each of the three arms, and not a benchmark family in the sense of Table~\ref{tab:breadth}. In all, \FactGymFamiliesWord{} families were attempted, with seeds 42, 43 and 44 in each arm on every family but Square, where the controller's seed 44 was never launched: eight of Square's nine runs were attempted and none completed, so \FactGymCompleteWord{} families completed. Two policy classes are represented and each family runs under exactly one of them, seven under ReFlow and two, Kitchen mixed and Kitchen partial, under ShortCut; the classes are not crossed with the families and no family is run twice. Eight completed in all three arms and are included in every count below. The owner record of every run, its last iteration, its final return, its within-evaluation standard deviation and its collapse count is \texttt{reinflow\_gym\_table.md}, and the cells below are read from it rather than recomputed.

The collapse level plotted in Figure~\ref{fig:gymstate} and Figure~\ref{fig:gymimage} is the \emph{Ever} definition on this benchmark: a run counts as collapsed if any plotted evaluation falls below half of the family's common tare, the mean iteration-zero return of its nine runs, whose run configurations all name the same base checkpoint; that is what the owner table reports as a count of evaluations below the level. In both figures the seed band is drawn only where every contributing seed has a value, so its width never comes from a seed having stopped earlier; a dashed mean means fewer than three seeds contributed and a hollow final marker means the runs behind it did not all reach the target iteration.

\textbf{Two rulers again, and they are not the LIBERO ones.} The quantity here is an episode return rather than a success count, so there is no Wilson interval to give. The two rulers are the spread of the three seeds' final returns and the largest within-evaluation standard deviation in the family. The rule of the owner record has two conditions and asks for both: the arm's mean lies outside the range of the other two arms' seeds, and its difference from the fixed arm exceeds the larger of the two rulers. Where either fails, the family is reported as within and no verdict is drawn for that arm on that task, which is the same rule this paper applies to its own instrument and the same reason it gives two numbers for its rescoring floor. This is a screening threshold against two measured spreads and not a hypothesis test: with three seeds per arm neither ruler carries a sampling distribution, exceeding both is not a claim of significance, and falling inside either is not evidence that two arms are equivalent.

\textbf{What the arms do.} No learned-against-fixed comparison exceeds both thresholds on any of the eight complete families, so the screening rule fires for that pair nowhere on this benchmark. That is a statement about resolution and not an equivalence, and Can shows why the difference matters: there the learned arm sits \num{+4.60} from the fixed arm, which does exceed \num{4.10}, the largest seed SD among the three arms (the controller's), and what it does not approach is the largest within-evaluation spread of \num{41}. The rule asks for both, so it is failed on one condition rather than on neither. The controller's own comparison separates on two of the eight under both readings, and both are downward. Under the owner's rule: Hopper, a difference of \num{-332.55} against rulers of \num{215.17} and \num{238.21}, and Kitchen mixed, \num{-2.24} against \num{0.23} and \num{0.53}. Under the stricter reading set out below, which governs the count, the controller's mean on Hopper lies \num{263.52} below the nearest reference seed, against a population seed ruler of \num{175.68} and the evaluation ruler of \num{238.21}, a margin of \num{25.31}; on Kitchen mixed it lies \num{2.24} below the nearest reference seed, against \num{0.19} and \num{0.53}. Kitchen mixed is the one of those two that is not a single continuous skill: it chains several manipulation subtasks within one episode. A candidate explanation for the drop there is that the gate's novelty and competence signals do not match the exploration a given stage of the chain needs, so that the scale is raised where acquired behaviour is disturbed or held down where exploration would still help. Stage-level behaviour was not analyzed in these runs, so this is a hypothesis about the negative result and not a measured cause, and no reading here depends on it. Among the remaining six families, the owner's rule does not identify a separation on Ant, Can, Humanoid, Kitchen partial or Walker. Kitchen complete meets both conditions of the owner's rule, but not the stricter reading below. Neither outcome establishes equivalence. On Can the controller sits \num{-6.72} from the fixed arm, which like the learned arm's \num{+4.60} exceeds that seed SD of \num{4.10} and does not approach the largest within-evaluation spread of \num{41}; the rule asks for both, so neither comparison is reported as a separation on Can, and neither is reported as a coincidence. On Kitchen complete the owner's rule identifies an upward separation: the controller's mean lies outside the other arms' seeds and its difference of \num{+1.44} exceeds the larger ruler, \num{1.20} against \num{0.80}, and the owner record reports a separation there. What withholds it from the count in this paper is the stricter reading set out in the paragraph that follows; under the owner's rule the family separates, and under the stricter reading it does not. Both readings are reported; the summary count is the stricter reading's. No family on this benchmark is reported in the controller's favour.

 \textbf{Why Kitchen complete is not counted.} Under the owner's rule Kitchen complete separates upward: the controller pools at \num{3.56}, outside the other arms' seeds, which span \num{1.00} to \num{3.38}, and its difference of \num{+1.44} exceeds the larger ruler, \num{1.20} against \num{0.80}; the owner record reports it that way. The reading that governs this appendix is stricter: the arm's mean must clear the nearest individual seed of the reference arms by more than the larger ruler, not only lie beyond it. Here the margin is \num{0.18} against a ruler of \num{0.98}, so the family is within. The two readings compute the seed ruler from the same three fixed-arm returns in two ways: the owner's rule takes their sample standard deviation, \num{1.20}, and the stricter reading their population standard deviation, \num{0.98}; the margin of \num{0.18} is inside under either. What makes the two readings differ on this family is measured: one fixed-arm seed sits at \num{1.00} with \num{30} of its \num{31} evaluations below the collapse line, which pulls the reference mean down without widening the seeds' span above it. The stricter reading was defined after the owner's verdicts were seen, and its scope is stated so it cannot be mistaken for the owner's rule: the shipped \texttt{reinflow\_family\_verdict.py} applies it to every complete family and prints it as the governing reading, and it agrees with the owner's rule on seven of the eight and departs from it on Kitchen complete alone, where it withholds a separation. Both results are reported, with the stricter one determining the summary count.

\textbf{Checkpoint availability.} Results in this appendix are computed from archived evaluation records. Final checkpoints were retained and verified for \texttt{can\_F\_s43} and \texttt{can\_R\_s44}. Checkpoints for the remaining runs were not retained, preventing additional policy evaluations and analyses requiring those original weights.

\textbf{What did not run, and what was never attempted.} Square was attempted and not completed: of its eight runs none reached the target iteration, so no arm has a single complete seed on it and every Square run is excluded from every count in this appendix. A family read from partial runs would be the completeness rule of Section~\ref{sec:setup} broken on a different benchmark. Transport, the third Robomimic task in the reference's own profile, was never attempted and has no row anywhere in the owner record. Neither absence is filled in and neither is carried as a pending cell.

\textbf{Why the baseline columns of Table~\ref{tab:reinflow} cannot be widened in this re-run.} The reference reports DPPO and FQL as its two headline baselines, and a per-task number for either would let this paper put a published comparison beside its own arms. Those numbers are not in the paper: the comparison is made in figures, and a search of the pinned repository at \texttt{e722e15} for a returns file or table found none. Reading a value off a plotted curve by eye was declined. The columns therefore stand empty as a recorded measurement, and the baseline breadth of this paper is unchanged by the re-run.

\begin{figure}[p]
\centering
\includegraphics[width=\textwidth]{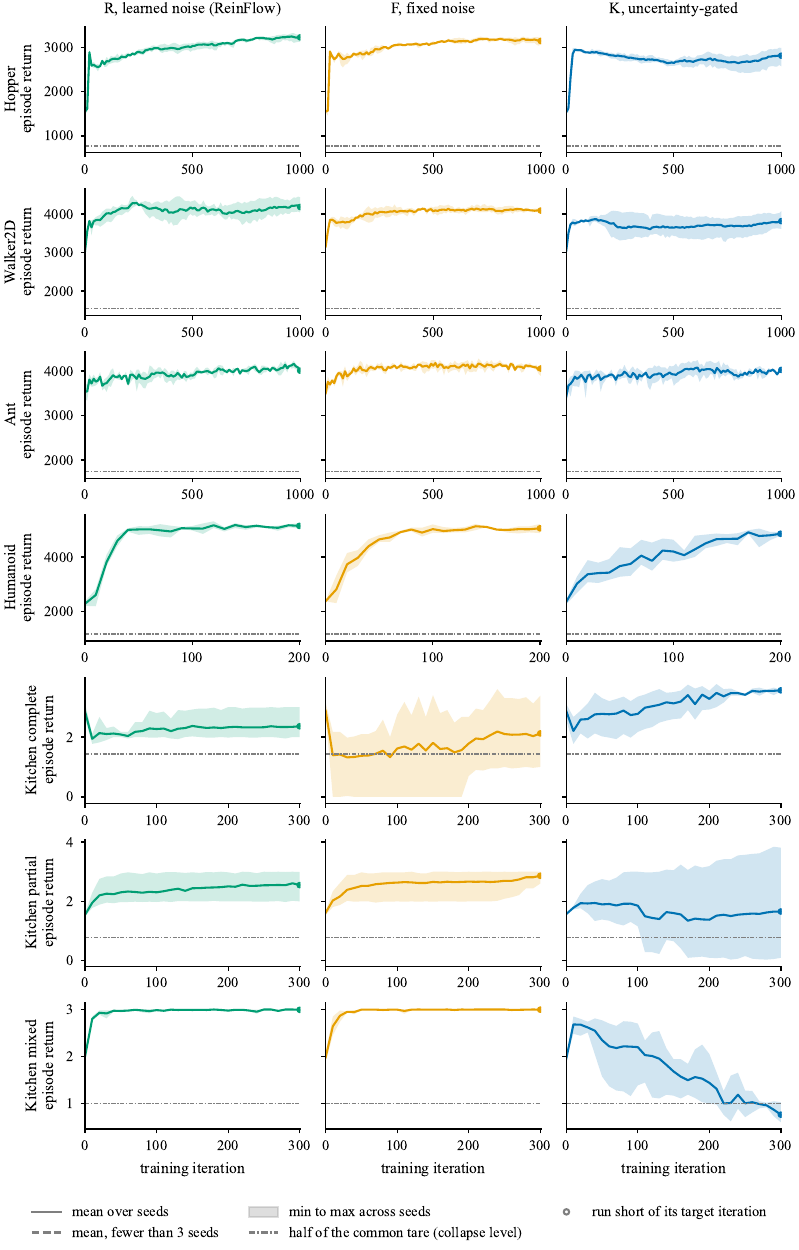}
\end{figure}

\begin{figure}[t]
\caption{Training curves on the state-based families, on the preceding page. Columns are the three arms: R runs the learned noise unchanged, F locks the noise to a constant, K gates it on an uncertainty signal. Within a family the run configurations of all three arms name the same base policy checkpoint, so the iteration-zero evaluation is a common tare and the dotted line marks half of it, which is this benchmark's collapse level and not the pooled success definition of the LIBERO campaign. F and K additionally run the learned noise for their first eight iterations and take their constant from it, diverging at iteration 8. This warm-up does not make F and K share a value: each run measures its own constant from its own first eight iterations, so the arms still differ in one thing, whether that constant is then held fixed or gated. It is also a different mechanism from the LIBERO campaign's critic warm-up, which is likewise eight iterations and holds off the competence signal. Each cell shows the mean over seeds with a band giving the minimum and maximum across them; the band is not a standard deviation. Produced by \texttt{make\_fig\_reinflow.py}.}
\label{fig:gymstate}
\end{figure}

\begin{figure}[!ht]
\centering
\includegraphics[width=\textwidth]{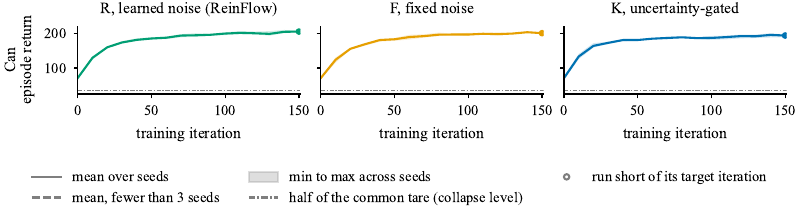}
\caption{Training curves on Can, the image-based family, in the layout of Figure~\ref{fig:gymstate} and with the same mean-and-band construction. Square is attempted but not completed, so it carries no curve and no number anywhere in this paper: of its eight runs none reached the target iteration. A dashed mean means fewer than three seeds contributed and a hollow final marker means the runs behind it did not all reach the target iteration. Produced by \texttt{make\_fig\_reinflow.py}.}
\label{fig:gymimage}
\end{figure}

\fi

\clearpage
\section{Measurement protocol, instruments and reproducibility}
\label{app:repro}
\textbf{The published figure this paper compares its tare against is not a state-of-the-art claim.} It is the LIBERO-Long row for the same 450M policy at the same size \citep{smolvla}. State-of-the-art numbers on this suite are set by models of other sizes, and nothing in this paper is measured against them; the comparison here bounds how much room sits above the tare our own runs start from, and it is that alone.

\textbf{The precision regime, read from the reference configuration.} These are the file and line numbers that Appendix~\ref{app:precision} refers to. The last row is ours and is read from the checkpoint rather than from a configuration file.

\begin{table}[h]
\centering
\small
\setlength{\tabcolsep}{4pt}
\begin{tabular}{>{\raggedright\arraybackslash}p{0.34\textwidth}l>{\raggedright\arraybackslash}p{0.40\textwidth}}
\toprule
file & line & what it sets \\
\midrule
\texttt{\footnotesize examples/\allowbreak embodiment/\allowbreak config/\allowbreak model/\allowbreak pi0.\allowbreak yaml} & 6 & \texttt{\footnotesize precision: null} \\
\texttt{\footnotesize examples/\allowbreak embodiment/\allowbreak config/\allowbreak model/\allowbreak pi0\_\allowbreak 5.\allowbreak yaml} & 6 & \texttt{\footnotesize precision: null, with the comment that it must not be changed} \\
\texttt{\footnotesize examples/\allowbreak embodiment/\allowbreak config/\allowbreak libero\_\allowbreak 10\_\allowbreak ppo\_\allowbreak openpi.\allowbreak yaml} & 118 & \texttt{\footnotesize precision: \$\{actor.\allowbreak model.\allowbreak precision\}} \\
\texttt{\footnotesize examples/\allowbreak embodiment/\allowbreak config/\allowbreak libero\_\allowbreak 10\_\allowbreak ppo\_\allowbreak openpi.\allowbreak yaml} & 152--154 & \texttt{\footnotesize param\_\allowbreak dtype, reduce\_\allowbreak dtype and buffer\_\allowbreak dtype all bound to the same variable} \\
\texttt{\footnotesize rlinf/\allowbreak config.\allowbreak py} & 379--384 & \texttt{\footnotesize use\_\allowbreak fsdp\_\allowbreak mixed\_\allowbreak precision = not (all\_\allowbreak none or all\_\allowbreak fp32)} \\
\texttt{\footnotesize examples/\allowbreak embodiment/\allowbreak config/\allowbreak training\_\allowbreak backend/\allowbreak fsdp.\allowbreak yaml} & 22--31 & \texttt{\footnotesize mixed\_\allowbreak precision null, amp\_\allowbreak autocast.\allowbreak enabled False, grad\_\allowbreak scaler.\allowbreak enabled False} \\
\texttt{\footnotesize rlinf/\allowbreak hybrid\_\allowbreak engines/\allowbreak fsdp/\allowbreak fsdp\_\allowbreak model\_\allowbreak manager.\allowbreak py} & 114, 533 & \texttt{\footnotesize nullcontext() when AMP is off; AdamW built directly on the model parameters} \\
\texttt{\footnotesize rlinf/\allowbreak models/\allowbreak embodiment/\allowbreak openpi/\allowbreak \_\allowbreak \_\allowbreak init\_\allowbreak \_\allowbreak .\allowbreak py} & 89 & \texttt{\footnotesize to\_\allowbreak bfloat16\_\allowbreak for\_\allowbreak selected\_\allowbreak params("bfloat16")} \\
\texttt{\footnotesize openpi, PyTorch path} & --- & \texttt{\footnotesize pytorch\_\allowbreak training\_\allowbreak precision defaults to bfloat16, no autocast, no grad scaler} \\
\texttt{\footnotesize lerobot/\allowbreak configs/\allowbreak policies.\allowbreak py} & 65 & \texttt{\footnotesize use\_\allowbreak amp: bool = False} \\
\texttt{\footnotesize lerobot/\allowbreak .\allowbreak .\allowbreak .\allowbreak /\allowbreak smolvlm\_\allowbreak with\_\allowbreak expert.\allowbreak py} & 80 & \texttt{\footnotesize VLM loaded with torch\_\allowbreak dtype="bfloat16"} \\
\texttt{\footnotesize ours, read from \texttt{iter\_\allowbreak 000036.\allowbreak pt}} & --- & \texttt{\footnotesize 1204 bfloat16 tensors against 57 fp32; optimizer moments bfloat16 in 253 of 284; no fp32 master copy} \\
\bottomrule
\end{tabular}
\caption{The precision lines of the pinned RLinf clone (commit \texttt{832db3f5}), of openpi and of LeRobot SmolVLA, and our own regime read from the weights of \texttt{k\_corrected\_s1337}, the run of Table~\ref{tab:round}. Autocast and the gradient scaler are off in the reference recipe, so training is bfloat16 in place rather than mixed precision. Supplementary file \texttt{reference\_precision.md}.}
\label{tab:refcfg}
\end{table}

\textbf{Instruments and tares.} All campaign arms run on identical rented RTX 3090 nodes (one arm per node) with an identical software image and a single trainer per GPU. The BC checkpoint scored on the campaign instrument gives \num{136/210} (0.648); the same weights scored on our local development GPU give \num{121/210} (0.576), a difference of 15 episodes with no change of weights or protocol. Every arm is therefore compared only to the tare measured on its own instrument.

\textbf{Episode identity is not preserved across evaluation geometry.} Scoring the same BC checkpoint on the same 210 initial-state identities with 21 and with 14 parallel environments returns the same total, \num{136/210} in both, while \num{56} of the 210 cells disagree. The shared ruler is therefore the total, not the individual episode. This bounds every cell-level comparison in this paper, including McNemar tests, whose paired cells are only meaningful between scans of equal geometry.

\textbf{Design principles for indefinite training.} Several choices that a benchmark paper would make differently were made for a policy that never stops training. There are no decay schedules: under continuous operation a decaying noise or learning rate reaches zero and learning ends, and resetting a schedule from a high value is arbitrary. There is no KL anchor to an earlier policy: under sustained drift the anchor goes stale, and its effect reduces to a learning-rate cut that can be obtained more directly; we accepted a more fragile policy for a single-variable comparison. Evaluation is always stochastic: a policy trained and deployed in SDE mode is scored in SDE mode, and the run-to-run spread that this produces is measured and reported. The instrument itself is part of the result: the same weights score differently on two GPUs, and a fleet that trains locally will need bit-level stability of its training and evaluation paths before local updates can be trusted to agree.

The limitations of the main text at full length, with their evidence:
\begin{itemize}
\item \textbf{What the mechanism is for, and what is not shown.} The target of a task-adaptive
noise scale is a policy that keeps learning without losing what it already does, and resistance to
per-task collapse under continued training is a necessary condition for that. Collapse is the
quantity we measure; the condition asks for its absence. It is not a sufficient condition either.
Our regime is simultaneous multi-task fine-tuning, not the sequential regime in
which lifelong learning is usually posed, so this paper measures a necessary property in the easier
setting and leaves the sufficiency claim to future work in the sequential one.
\item \textbf{No gain over behavior cloning.} No arm exceeds its own tare; what is measured is the
presence, timing and seed-to-seed distribution of collapse, not absolute success. Several candidate
explanations remain unresolved in this design, and none substitutes for another. The policy is a 450M-parameter model, and the published LIBERO-Long figure for that
size, \num{71} of 100 trials \citep{smolvla}, is above both of our tares: \num{136/210}
(0.648) on the campaign instrument and \num{121/210} (0.576) on the local one. The gaps are \num{0.0624} and \num{0.1338}, about \num{1.66} and \num{6.75} times each instrument's rescoring SD, and the Wilson
intervals overlap in both cases. Read instead against the published figure's own interval, \num{0.6146} to \num{0.7899} for \num{71} of 100 with a half-width of \num{0.0876}, the campaign gap sits inside it, and the local gap, at \num{1.53} times that half-width, does not. The starting point is therefore near the published figure for
its own size rather than far below it. The protocols also differ: ten trials on each of ten
tasks against thirty episodes on each of seven, at $\sigma_{\mathrm{eval}} = 0.25$ on the
480-step ruler. LIBERO-10 is the hardest of the four suites,
with long-horizon multi-stage tasks. What is measured here is narrower than a ceiling: the surface
measurement shows gradient reaching the action expert, with 112 of its 145 tensors changing in at
least one element, which
rules out the one plumbing fault that measurement tests and leaves the remaining candidate explanations unresolved.
Whether the policy's capacity is the binding constraint was not measured. The campaign also ran on
a small fraction of the reference recipe's update budget, and no arm in it was given the budget at
which the reference reports its figure.
\item \textbf{Precision.} Training ran in bfloat16 in place with no fp32 master copy and with autocast and the gradient scaler both off (Table~\ref{tab:refcfg}), which
rounds away part of every small update. We did not deviate here, we followed the reference: all
three reference recipes keep no master copy. What we add is that the property was measured rather
than assumed. Over three outer iterations, 96.02\% of the elements in the action-expert body stayed
bit-identical while 112 of 145 tensors moved in at least one element. That measurement shows the
swallowing is possible, not that it caused the outcome. Separating the two is what the fp32 pilot is
for, and both of its arms are measured to iteration 200 and read in Appendix~\ref{app:precision}.
\item \textbf{Learning rate under fp32.} The fp32 master-copy pilot holds the learning rate
at the reference value and uses the same seed and the same pod period as its bfloat16 twin. It is
not a controlled single-variable comparison. For the controller pair, each run's
\texttt{run\_config.json} was written when its process was resumed rather than when it started;
\texttt{diff\_run\_config.py} compares 104 fields of the two and finds three that differ, the fp32
master flag and the run and checkpoint-archive directory names, and the two processes ran from
different code commits, which that comparison does not cover. Both files are in the archive as
\texttt{run\_config\_k\_s1337\_bf16.json} and \texttt{run\_config\_k\_s1337\_fp32.json}. For the
fixed pair no \texttt{run\_config.json} is on record, so those two configurations cannot be
compared field by field from the records that exist. The pilot is reported as an observation about
two regimes throughout. This is separate from the
matched comparison of the three exploration arms in the main bfloat16 campaign, which is
single-variable by design and is where the paper's result is measured. Because the pilot's own reading
shows that removing the rounding changes how much of each update survives, the reference rate
is not necessarily the right rate under fp32, and retuning it is outside this work. We report the observed contrast at the reference learning rate; an optimized fp32 recipe was not evaluated.
\item \textbf{Single model, single benchmark.} The findings are measured on one policy and
one suite, a seven-task pool of LIBERO-10, and do not generalize to another model or another suite.
Three seeds per arm, all measured to iteration 200; one controller seed was continued from its iteration-180 checkpoint to get there.
\item \textbf{The learned-noise family is represented by one member.} The Flow-Noise variant of
\citep{pirl}, which learns the per-step noise inside a discrete-time MDP over the denoising steps
and therefore obtains exact log-likelihoods rather than the SDE approximation, was not run. Our R arm
should be read as one member of that family and not as its best case.
\item \textbf{Evaluation budget.} Canonical scans use 30 episodes per task, below the de facto
LIBERO standard of 50, so every scan rests on 30 episodes per task and a pooled plateau on 150.
The repeat scans below measure the rescoring noise and enter no point and no pooled window. Collapse thresholds at one half of a low tare sit
close to the rescoring noise floor, which is why the pooled and sustained definitions accompany the
point-wise one.
\item \textbf{Instrument effects.} The same weights, rescored under the same protocol with only
the evaluation seed changed, show checkpoint-dependent variation across repeated evaluations: run-to-run standard
deviation is 0.0375 on the campaign instrument over four scans and 0.0198 on the local tare over
three, a figure that belongs to those weights and not to that instrument, which reaches \num{0.0676} at a degraded checkpoint (\emph{Rescoring noise and policy health}, below),
and the campaign figure is itself one of two measurements of the same quantity: four
scans of the controller's seed-3337 iteration-200 checkpoint give \num{0.0375} (\num{127},
\num{132}, \num{118}, \num{115} of \num{210}); those four are not one directory, three of them are the repeat set and the fourth is the series' own canonical scan, and the three repeats alone give \num{0.0432}, so a reader who finds the repeat set and stops there recovers a different constant from the one this paper carries. The same four-scan design on seed 1337's checkpoint gives
\num{0.0147} (\num{116}, \num{115}, \num{122}, \num{118}). Six repeat sets on this instrument span
\num{0.0048} to \num{0.0375}. We carry the largest everywhere, and the direction that choice
leans is not the same in both kinds of statement: where we call a difference real it yields the
smallest multiple and is the conservative reading, and where we call a difference noise it is
the easiest test to pass and is the favourable one. The single claim of the second kind is the
spread of the three controller seeds, and Section~\ref{sec:results} gives both readings of it
rather than the one that suits us. On the local instrument, the rescoring study described below under \emph{Rescoring noise and policy health} gives \num{0.0619} and \num{0.0676}, which are \FactPairedPctLate{}\% and \FactPairedPctEarly{}\% of the Wilson half-width
of a single 210-episode scan taken at each set's own pooled rate, so there neither ruler contains the other. Every comparison is read within one instrument, and local and
campaign numbers are never placed in the same column.
\item \textbf{Scope of the evaluated controller.} The reported effect is obtained with two signal
families and a four-cell rule. Additional drivers present in the code were inactive, so the results
apply to the evaluated rule. The rule in
\eqref{eq:novel}--\eqref{eq:sigma} is driven by those two signals and that four-cell table, and it departs
in two places from the fuller design it came from: the familiar-and-low-competence cell multiplies
by 1.2, raising $\sigma$ to 0.36 above the 0.30 base, and two drivers are switched off. The design
gives that cell two rationales (Appendix~\ref{app:controller}): the four-state table of the design
notes lowers the scale there, and the sixteen-cell variant sets 0.7 when its other two signals are
calm, 1.0 when one of them fires and 1.3 when both do, while the literature synthesis behind the
specification asks for a moderate scale, which the running rule follows. This is a scope statement, not a defect report;
the arms ran with this rule, and the effects reported here belong to it and do not generalize to the
fuller design, which was not run.
\item \textbf{Exposure.} Every arm of the LIBERO campaign reached iteration 200 and was
scanned every 20 iterations, and that is the scope of the claim: the local-instrument row of
Table~\ref{tab:pool} pools a different window, and the re-run of Appendix~\ref{app:reinflowgym}
runs on per-task horizons and is not scanned on this cadence at all. One controller seed reached
200 after being resumed from its own iteration-180 checkpoint rather than running without
interruption, and the effect of that resumption on its reading was not measured. Campaign runs
that did not reach the target iteration are excluded from every seed-level count in this paper and
stay in the released measurement files.
\end{itemize}

Per outer iteration every LIBERO arm, campaign and local alike, collects $14$ environments $\times\, 73$ rollout steps $\times\, 2$ accumulation rounds $= 2044$ samples and takes four optimizer steps, one per PPO epoch, each over the whole buffer. Both figures are read from the run's own counter rather than computed: the training loop prints \texttt{optimizer\_steps\_this\_outer=4 (expected 4) samples\_per\_step=2044} from the advantage tensor itself, and the trainer asserts the printed count against the expected one before the iteration closes (archive \texttt{manifests/banner\_k\_s1337\_local\_env14.log}, and the \texttt{measured\_OBSERVED} block of \texttt{configs/run\_config\_k\_s1337\_bf16.json} and its fp32 twin). Actor learning rate is $5\times10^{-6}$ and critic learning rate $10^{-4}$; both are inside the ranges the reference recipe uses for LIBERO (actor $10^{-5}$ to $5\times10^{-6}$, critic $10^{-4}$), and our samples per iteration, 2044, are within four samples of the reference batch size of 2048. The trainable surface is the full action expert (98.2M parameters in the body and 1.6M in the flow head), LoRA adapters on the vision-language backbone (0.8M) and the value head (0.4M), 101.1M trainable parameters against 350.2M frozen and 451.3M in all, the groups and counts of Table~\ref{tab:round}.

We use seven of its ten tasks, \texttt{libero\_10:\{1,2,3,5,6,7,9\}}, read from the verified launch arguments of the campaign runs. Throughout the paper t0--t6 name these seven in this order, so t0 is \texttt{libero\_10:1} and t6 is \texttt{libero\_10:9}; an index written with the \texttt{libero\_10:} prefix is a LIBERO index and never a pool label. The three excluded tasks, \texttt{libero\_10:\{0,4,8\}}, were dropped because the BC checkpoint's success on them was at or near zero on our instrument at the budget in force when the pool was fixed (the local tare on \texttt{libero\_10:0} reads 0.002 in the project record), and the pool was chosen so that every task had room both to fall and to rise under the instrument this study can afford: one suite, LIBERO-10, scored at 30 episodes per task per scan, on which a task starting at or near zero leaves a fall with nothing to measure and a rise with one episode of resolution. The limit is the budget and the instrument, not the tasks: the record notes that at a longer horizon some of those three begin to succeed, so they were not excluded as impossible. The exclusion was made on the 300-step budget, and the tares of those three tasks on the 480-step ruler were not measured. Success follows the LIBERO protocol \citep{libero} (goal predicate satisfied for at least ten consecutive steps) at a horizon of 480 steps.

Per iteration this is parity with the reference recipe's LIBERO batch of 2048; the gap is in the number of iterations and therefore in total updates: our arms take 800 optimizer steps in total, four per outer iteration over 200 iterations.

\textbf{Canonical evaluation.} Every 20 iterations, and at the end of training, a canonical scan of 210 episodes (30 per task, fixed initial-state identities) in SDE mode with $\sigma_{\mathrm{eval}}=0.25$. For six runs on the campaign instrument the final weights were scored three or four times with only the evaluation seed changed, and six more such sets were taken on the local instrument at six checkpoints; these repeat sets measure the rescoring noise and are not pooled. All twelve sets are in the supplementary archive.

\textbf{Rescoring noise and policy health.} On the local instrument, rescoring the controller seed-1337 weights gives a run-to-run SD of \num{0.0619} at iteration 200 and \num{0.0676} at iteration 100 (three scans each, evaluation seed the only difference), against \num{0.0198} for the healthy BC weights on that same instrument, ratios of \num{3.12} and \num{3.41}. This comparison is bound to the local instrument: the campaign run-to-run SD of \num{0.0375} is a different ruler and the two are not combined into one ratio. No degraded checkpoint was repeated on the campaign instrument, so this paragraph does not generalize beyond the local instrument and is a single-arm observation.

\begin{table}[t]
\centering
\scriptsize
\caption{The measurement files in the supplementary archive, what each holds, and how each was made.}
\label{tab:supplementary}
\begin{tabular}{@{}>{\raggedright\arraybackslash}p{0.21\textwidth}>{\raggedright\arraybackslash}p{0.42\textwidth}>{\raggedright\arraybackslash}p{0.31\textwidth}@{}}
\toprule
file & what it holds & how it is made \\
\midrule
\texttt{arm\_\allowbreak{}table.\allowbreak{}md} & every arm, its pooled value, both rulers, the four collapse columns & \texttt{code/\allowbreak{}make\_\allowbreak{}arm\_\allowbreak{}table.\allowbreak{}py} \\
\texttt{t1\_\allowbreak{}plateau\_\allowbreak{}pool.\allowbreak{}md} & the plateau pool of every series & \texttt{code/\allowbreak{}make\_\allowbreak{}figures\_\allowbreak{}legacy.\allowbreak{}py} \\
\texttt{t5\_\allowbreak{}collapse\_\allowbreak{}definitions.\allowbreak{}md} & the collapse definitions read side by side & \texttt{code/\allowbreak{}collapse\_\allowbreak{}three\_\allowbreak{}ways.\allowbreak{}py} \\
\texttt{t6\_\allowbreak{}fp32\_\allowbreak{}pilot.\allowbreak{}md} & the fp32 master-copy pilot & \texttt{code/\allowbreak{}make\_\allowbreak{}t6\_\allowbreak{}reading.\allowbreak{}py} \\
\texttt{parameter\_\allowbreak{}movement\_\allowbreak{}by\_\allowbreak{}arm.\allowbreak{}md} & relative movement of each trainable group per arm and window & \texttt{code/\allowbreak{}measure\_\allowbreak{}param\_\allowbreak{}movement.\allowbreak{}py} \\
\texttt{bf16\_\allowbreak{}swallowed\_\allowbreak{}updates.\allowbreak{}md} & the element-level bfloat16 measurement and the fp32 arithmetic beside it & authored record, append-only journal of readings taken from checkpoint tensors \\
\texttt{reference\_\allowbreak{}precision.\allowbreak{}md} & the precision lines of the reference recipes, read from their code & authored record, append-only journal of file and line references \\
\texttt{prereg\_\allowbreak{}fp32\_\allowbreak{}pilot.\allowbreak{}md} & the fp32 pilot's pre-registration & authored record, written before any result was seen \\
\texttt{prereg\_\allowbreak{}sigma\_\allowbreak{}ablation.\allowbreak{}md} & the sigma ablation's pre-registration & authored record, written before any result was seen \\
\texttt{ace\_\allowbreak{}calibration.\allowbreak{}md} & the ACE signal's calibration & \texttt{code/\allowbreak{}ace\_\allowbreak{}calibration.\allowbreak{}py} \\
\texttt{reinflow\_\allowbreak{}gym\_\allowbreak{}table.\allowbreak{}md} & every gym run, its last iteration, final return, within-evaluation SD and collapse count & \texttt{code/\allowbreak{}gym\_\allowbreak{}table.\allowbreak{}py} \\
\bottomrule
\end{tabular}
\end{table}

\clearpage
\section{Anticipated questions and clarifications}
\label{app:faq}
The questions below clarify the experimental scope and the interpretation of the results,
with the measurement each answer rests on.

\textbf{Why a single benchmark suite?} We selected LIBERO-10 because its measured starting success left room to examine both deterioration and improvement within our evaluation budget. The behavior-cloning policy every arm starts from scores \num{0.9444}
on Object, \num{0.8611} on Spatial and \num{0.8333} on Goal, against
\num{0.5762} on LIBERO-10, so on three of the four suites a task starts near the ceiling, with
almost no room above it and its collapse threshold, one half of its own tare, far below where it
starts. The suite this paper needs is the one with room in both directions, because both erosion
and protection are being asked about. This is the same rule that selected the
seven-task pool inside LIBERO-10, where a task whose tare is at zero leaves nothing to protect and
a task at the ceiling leaves nothing to gain. One scope note travels with those numbers. The three saturated suites
were scored on the 300-step ruler at 18 episodes per task and only over four of their ten tasks,
while LIBERO-10 is scored on the 480-step SDE ruler at 30 episodes per task over seven, so the
columns are read beside each other and not differenced.

\label{app:breadth}
\begin{table}[h]
\caption{Evaluation profiles of two recent RL fine-tuning papers, DPPO \citep{dppo} and WSRL \citep{wsrl}, and this work. Counts are read from the full text of arXiv version 3 of each paper, from its experiment sections and appendices; a policy-class cell counts the classes the method itself fine-tunes, and the classes fine-tuned only as baselines are named after it. The first row counts benchmark families, the unit of Table~\ref{tab:reinflow}: a benchmark family is a suite such as OpenAI Gym or Robomimic, not one task inside it, and DPPO's and WSRL's counts are suites. Ours are LIBERO-10 and the benchmark families of Table~\ref{tab:reinflow} that hold a task of Appendix~\ref{app:reinflowgym} completed in all three arms, \FactBenchmarkFamilies{} in all. Our two policy classes are the pretrained vision-language-action policy of the main campaign and the small flow policies of that re-run.}
\label{tab:breadth}
\centering\footnotesize
\resizebox{\textwidth}{!}{%
\begin{tabular}{lccc}
\toprule
 & DPPO & WSRL & this paper \\
\midrule
benchmark families & 5 & 4 & \FactBenchmarkFamilies \\
policy classes & 1, diffusion (MLP, UNet, ViT-MLP); Gaussian and GMM as baselines & 1, soft actor-critic policy (MLP) & \FactPolicyClasses \\
real robot & yes, Franka, 20 trials & yes, Franka peg insertion, 20 trials & no \\
named baselines & 11 & 6 in simulation, SERL on the robot & none; \FactArms{} arms against the policy's own BC tare \\
seeds & 5 (Gym, D3IL), 3 (Robomimic, Kitchen, Furniture-Bench) & not stated & \FactSeeds{} per arm \\
variability reported & mean over seeds; two figures state SD not shown & no measure stated & Wilson per scoring and repeat-scoring SD \\
\bottomrule
\end{tabular}
}
\end{table}

\subsection{What this paper measures that the learned-noise reference does not}
\label{app:beyond}

The reference for our learned-noise arm \citep{reinflow} fine-tunes flow-matching policies with online RL, which is the mechanism we port. Three things that our setting adds are absent from it, and we state them as differences in scope rather than as shortcomings, since two of the three are named by its own authors as future work.

\begin{enumerate}
\item \textbf{The policy is not a VLA there and is here.} Its experiments are flow-matching policies over state and pixel input for continuous control. Its own conclusion says so and names the gap: ``our current experiments use relatively small networks, and scaling ReinFlow to large flow-based vision-language-action (VLA) models remains an exciting challenge. We leave these to future work'' \citep[\S7]{reinflow}. Our policy is a 450M-parameter pretrained VLA, so the rounding property we measure in Appendix~\ref{app:precision} is a property of that regime and could not have appeared in theirs.
\item \textbf{The policy is not language-conditioned there and is here.} No task in that work carries a language instruction; the task identity is the environment. Every task in our suite is specified by an instruction to the same policy, which is what makes per-task interference a question about one network rather than about several.
\item \textbf{Per-task collapse is not measured there and is the object here.} Its Franka Kitchen tasks are described as long-horizon multitask planning, so the setting is multi-task; what is absent is the measurement. The words collapse, forgetting, interference and per-task do not occur in its text, and its reported quantity is an aggregate episode reward or success rate per task-suite. Our four collapse definitions, the survival analysis and the repeat-scoring ruler exist because an aggregate that is flat can hide tasks going to zero, which is the failure we report.
\end{enumerate}

The two evaluations lie on different axes. That work compares against two named baselines; this one names none and reads its arms against the policy's own behavior-cloning tare. In benchmark families that work has three and this one \FactBenchmarkFamiliesWord{}, LIBERO-10 and the three of Table~\ref{tab:reinflow}, which Appendix~\ref{app:reinflowgym} re-runs. This one measures a failure mode at the task level on a policy class that work does not cover.

\textbf{No arm beats behavior cloning. Why is this a paper?} Because the claim is about which
tasks survive continued updating, not about aggregate success, and in this regime the two
questions have different answers. The controller preserves every task under the pooled definition, although its aggregate
success remains below its behavior-cloning baseline. The absence of an aggregate gain is measured
rather than argued away. The update budget is a small
fraction of the reference recipe's, and training runs in bfloat16 with no fp32 master copy, under
which \num{96.02}\% of the action expert's elements are bit-identical across three
consecutive iterations while \num{112/145} of its tensors move in at least one element, with
the frozen side bit-identical in \num{100}\% of its elements as the null control that
shows the comparison works.

\textbf{Is the controller simply exploring less?} Not on the logged runs. The three controller runs with available applied-noise logs are the local run with seed 1337 and the campaign runs with seeds 1337 and 2337. Mean applied noise is \num{0.2399} on the local run over 144 outer iterations, \num{0.2444} on the campaign run with seed 1337 over all 200 and \num{0.2353} on the campaign run with seed 2337 over the 155 its log covers, against the fixed arm's \num{0.2500} with a standard deviation of \num{0.0000}, which is what a fixed scale means; the learned arm reads \num{0.2340} over 146 outers beginning at iteration 54. The controller therefore sits between \num{0.006} and \num{0.015} below the fixed arm depending on the run and on how much of it its log covers, and the difference is smaller than the per-task spread it produces on the same logs. The windows are of different lengths, one of the three campaign seeds logged no applied noise at all, and the learned arm's window excludes its early iterations, so this comparison is limited to the logged runs and their available windows. Lower fixed noise was run as the direct test: $\sigma = 0.18$ pools at
\num{0.496} with one collapsed task on its single seed, and $\sigma = 0.15$ pools at
\num{0.559} with none on seed 1337 and at \num{0.4629} \num{[0.4329, 0.4931]} with two,
t0 and t4, on seed 2337. Its volatility is measured rather than asserted: read at iterations 120, 160 and 200, each time over the five canonical points ending there, the 0.15 pooled value moved \num{0.610}, \num{0.577}, \num{0.559}, a drift of \num{0.051} larger than the campaign rescoring floor of \num{0.0375}, and it carries one sustained collapse on the task it lost at iteration 120 and has since recovered out of the pooled window. On a second measure, the SD of the five points inside the final window, the three controller seeds move less: \num{0.0123}, \num{0.0170} and \num{0.0124} for seeds 1337, 2337 and 3337 against \num{0.0364} for this ablation. The comparative statement is that a low fixed scale reduces collapse and is more volatile than the controller across the window measured. A lower scale therefore slows the decline without stopping it, and
the ordering is not what a pure noise-magnitude account predicts; but the second seed of the
0.15 ablation collapses two tasks, so a low fixed scale is not a substitute for the controller
and we do not present it as one.

\textbf{You report four collapse definitions. Which one is real?} All four, because they answer
different questions and their disagreement is itself a result. Two cases show why we do not
choose. Controller seed 3337 crosses on t4 at outer 160, recovers it at 180 and is below again
on its last scan at 200, so it counts under the last-scan and ever definitions and not under the
sustained one, which asks for two consecutive scans. Its pooled count is zero because the mean of
that task over the last five scans, \num{0.253}, stays above its threshold of \num{0.2}.
 Fixed seed 3337 is the mirror
image: it holds t0 below the threshold across the scans at outer 80, 100 and 120 and then
recovers it, so it counts as sustained while its pooled and last-scan counts are both zero.
\textbf{How much of this is evaluation noise?} Every comparison is given on two rulers, because
either one alone is half an answer. A single 210-episode scan carries a Wilson 95\% interval
of half-width about \num{0.067}. Rescoring the same weights with only the evaluation seed
changed gives a run-to-run standard deviation of \num{0.0375} on the campaign instrument
(four scans of one checkpoint: 0.6048, 0.6286, 0.5619, 0.5476) and \num{0.0198} on the local
instrument. Neither ruler contains the other; on the local instrument the run-to-run SD reaches
\FactPairedPctLate{} and \FactPairedPctEarly{} percent of the Wilson half-width a single scan
would carry at each checkpoint's own pooled rate, at the two controller checkpoints of the paired
study in Appendix~\ref{app:repro}, so
repeatability is reported separately from the interval for a single scan. Every figure this paper gives for this ruler comes from repeat scans
of one checkpoint and none from discordant-cell arithmetic, which measures a different thing.

\section{Extended discussion and outlook}
\label{app:outlook}

\textbf{What the mechanism is for.} The property measured here, not collapsing under continued updates, is a precondition for a policy that remains under online RL for its whole working life. The world we design for is a fleet in which every robot trains locally on the same base policy, keeps learning while it works, and exchanges experience rather than weights: model weights of tens of gigabytes cannot move to millions of robots after every update, and would not be worth moving, while a few megabytes of selected experience can. In that world a task that reaches zero measured success stops producing the reward that would teach it back, which is what makes the loss expensive rather than what makes it irreversible; this paper measures tasks that fell below the collapse line and later recovered, so zero on an instrument is not evidence that the competence is gone, and a policy that spends its capacity perfecting one task it already performs pays for it with the others. The controller studied here is the first piece of that design: a noise gate that spends exploration where the policy is uncertain and withholds it where the policy is competent, without any label that the world would not supply.

First, task collapse is the dominant phenomenon in this regime: fixed noise drifts downward at a seed-dependent rate and collapses tasks, the evaluated learned-noise arm collapses early and systematically, at a cost we do not quantify because update time was never instrumented. Second, the controller keeps the policy near its behavior-cloning level and loses no task under the pooled definition in any of its three seeds, and one task in one seed when single scans are read. The three controller seeds spread by a standard deviation of \num{0.013}, with the widest pair at \num{0.0248}, which is inside the rescoring floor measured on seed 3337's iteration-200 checkpoint (\num{0.0375}) and above the one measured on seed 1337's (\num{0.0147}); both are reported in Section~\ref{sec:results} and neither is treated as the floor. Third, the controller's mean applied noise sits between \num{0.006} and \num{0.015} below that of the fixed arm across the three logs that carry it, while the spread it produces across tasks is larger than that difference on the same logs, so the two arms differ in where the noise goes more than in how much of it there is; a lower fixed noise slows but does not stop the decline. Fourth, no arm improves on the behavior-cloning baseline, and two properties of the regime are measured beside that result without being separated from each other or from it: the small update budget and the bfloat16 update rounding inherited from the reference recipe. Fifth, we measure the rescoring noise of a single 210-episode evaluation and the difference between two evaluation instruments running the same weights, and show that neither is negligible; we therefore report every comparison on two rulers.

No arm improved on behavior cloning in this budget; the rounding property of the reference recipe and the size of the parameter displacements at which tasks collapse are findings in their own right, and neither is offered as the cause of that outcome. The results were obtained with the first version of the controller and came out close to what its design predicted. We take them as the first step of a longer program toward policies that remain under online RL for their whole working life, in which exploration, update size and plasticity are governed by the policy's own uncertainty rather than by hand.

\textbf{What comes next.} The second channel is the update itself. A person who has learned a skill stops actively learning it and does not lose it by using it; a policy that keeps a high learning rate on a task it already performs can degrade that task or overwrite others whose reward has not arrived for a while. A gate on the update size by the same signals is the next mechanism; managed plasticity, making room for new competence when capacity is exhausted, is the one after it, and it will need experiments at a scale where many tasks have been mastered. The novelty signal itself must be tested at scale: a single random-network detector may saturate over hundreds of tasks, and episodic forms of novelty are the candidate replacement. The full four-signal design, with the action-conditional-entropy signal and the value trend, was measured only through its logs here. Beyond the single policy, the same signals extend to the fleet. A central curator that trains nothing can filter the incoming experience, keep the fraction that is novel or that the policy fails on, and distribute that fraction to every robot; each robot trains locally on the same selected data from the same starting weights and stays in step with the others up to floating-point drift, and no model weight ever moves. Experience the policy already handles well carries little gradient and is not worth transmitting or training on; the uncertainty signals that gate exploration here would gate data selection there. We state this as the direction of the program rather than as a result. A calibrated estimate of success from the same signals would also let a single overseer watch a fleet and intervene before a failure, which we leave to future work.

\end{document}